\documentclass[sigconf]{acmart}
\copyrightyear{2018} 
\acmYear{2018} 
\setcopyright{acmcopyright}
\acmConference[MM '18]{2018 ACM Multimedia Conference}{October 22--26, 2018}{Seoul, Republic of Korea}
\acmBooktitle{MM '18: 2018 ACM Multimedia Conference, Oct. 22--26, 2018, Seoul, Republic of Korea}
\acmPrice{15.00}
\acmDOI{10.1145/3240508.3240652}
\acmISBN{978-1-4503-5665-7/18/10}

\usepackage{booktabs} 

\usepackage{graphicx}
\usepackage{amsmath,amssymb} 
\usepackage{color}
\usepackage{textcomp}
\usepackage{xspace}
\usepackage{multirow}
\usepackage{subcaption}
\usepackage{tabularx}
\usepackage{float}
\usepackage{xcolor}

\makeatletter
\DeclareRobustCommand\onedot{\futurelet\@let@token\@onedot}
\def\@onedot{\ifx\@let@token.\else.\null\fi\xspace}

\def\eg{\emph{e.g}\onedot} 
\def\ie{\emph{i.e}\onedot} 
 
\def\etc{\emph{etc}\onedot} 
 
\def\etal{\emph{et al}\onedot}

\editor{Jennifer B. Sartor}
\editor{Theo D'Hondt}
\editor{Wolfgang De Meuter}

\begin{document}
\title{ModaNet: A Large-scale Street Fashion Dataset with Polygon Annotations}

\author{Shuai Zheng}
\affiliation{%
  \institution{eBay Inc.}
  \streetaddress{2025 Hamilton Ave}
  \city{San Jose}
  \state{California}
  \postcode{95125}
}
\email{shuzheng@ebay.com}

\author{Fan Yang}
\affiliation{%
  \institution{eBay Inc.}
  \streetaddress{2025 Hamilton Ave}
  \city{San Jose}
  \state{California}
  \postcode{95125}
}
\email{fyang4@ebay.com}

\author{M. Hadi Kiapour}
\affiliation{%
  \institution{eBay Inc.}
  \streetaddress{199 Fremont St.}
  \city{San Francisco}
  \state{California}
  \postcode{94109}
}
\email{mkiapour@ebay.com}

\author{Robinson Piramuthu}
\affiliation{%
  \institution{eBay Inc.}
  \streetaddress{2025 Hamilton Ave}
  \city{San Francisco}
  \state{California}
  \postcode{95125}
}
\email{rpiramuthu@ebay.com}

\renewcommand{\shortauthors}{S. Zheng et al.}

\begin{abstract}
Understanding clothes from a single image would have huge commercial and cultural impacts on modern societies. However, this task remains a challenging computer vision problem due to wide variations in the appearance, style, brand and layering of clothing items. We present a  new database called ``ModaNet'', a large-scale collection of images based on Paperdoll dataset~\cite{DBLP:conf/iccv/YamaguchiKB13}. Our dataset provides $55,176$ street images, fully annotated with polygons on top of the 1 million weakly annotated street images in Paperdoll. ModaNet aims to provide a technical benchmark to fairly evaluate the progress of applying the latest computer vision techniques that rely on large data for fashion understanding. The rich annotation of the dataset allows to measure the performance of state-of-the-art algorithms for object detection, semantic segmentation and polygon prediction on street fashion images in detail.
\end{abstract}

\ccsdesc[500]{Computing methodologies~Computer vision}
\ccsdesc[500]{Computing methodologies~Image segmentation}
\ccsdesc[500]{Computing methodologies~Object detection}

\keywords{Fashion; Computer Vision; Dataset}

\maketitle

\section{Introduction}

Fashion understanding and analysis have been a popular topic in computer vision community for several years. 
With abundant visual information available, fashion is a natural domain to apply various computer vision techniques to improve online shopping experience and create significant business value.
Due to the enormous variety of clothing types and unpredictable clothing appearance, understanding fashion even in a single image remains an open problem.
Apart from identifying garments from a large number of categories and differentiating abstract styles from each other, which are already difficult, tasks such as localizing and segmenting fashion items are more desirable for real-life applications, including online shopping, personalized recommendation, and virtual try-on, \etc.
For example, given a street photo containing a celebrity, a shopper might be particularly interested in a specific fashion item, such as the pair of shoes the celebrity wears, and exploring more choices of the same style, shape or color. 
In this case, the shoes may only occupy a small portion of the entire photo, while the majority of the photo is not relevant to this shopper's intent.
Therefore, it is important to automatically localize and segment the shoes rather than sending the entire image to a visual search engine that may return irrelevant results.
By analyzing individual fashion items separately and the relationship between them, we move towards understanding the outfit as a whole.
In this context, automatic detection and segmentation is even more challenging as it requires more granular annotations for garment layers rather than image-level labels only, as well as other complications introduced by difficult body pose and clothing deformation. 

To address the above-mentioned problems, the first step is to have a high-quality, large-scale dataset to facilitate training models. 
Such fine-grained annotations are usually expensive to acquire as they require numerous human resources and domain knowledge compared to generating image-level labels. 
Previous works~\cite{DBLP:conf/iccv/YamaguchiKB13,DBLP:conf/cvpr/YamaguchiKOB12,DBLP:journals/tmm/LiuFDXHHY14} propose several datasets with pixel-level annotations for fashion parsing, but their annotated sets are limited to thousands of images only. 
Moreover, these datasets mainly focus on street fashion images containing people without many pose variations, hence the trained model may only work well in limited scenarios. 
Recently, Liu~\etal propose a large-scale dataset called DeepFashion. Apart from image-level labels specifying the type of clothes, the DeepFashion dataset also contains richer information such as binary attributes and landmarks that are useful to improve clothes retrieval performance. 
Although attributes and landmarks provide certain lower level information, they are still too coarse to train a good model to localize and segment individual fashion items.

In this work, we introduce a new fashion dataset called ModaNet consisting of $55,176$ high-quality, fully-annotated street fashion images on top of the 1 million weakly labeled images in the Paperdoll dataset~\cite{DBLP:conf/iccv/YamaguchiKB13}.
For the images that are fully-annotated, we have obtained pixel-level segmentation masks and polygons enclosing individual fashion items. 
We annotate 13 categories that are of the most interest in research and real-world commerce applications.  
Compared to its precedents, the ModaNet dataset has the following improvements. First, our ModaNet dataset is $10\times$ larger compared to other fashion datasets with pixel-level annotations, thus providing more data to train a better model specialized in detection and segmentation. 
Second, the annotated images are carefully selected to ensure diversity in human poses, \ie, not limited to the frontal view. 
In this way, training on this data grants a more generalizable model. Third, not only we provide pixel-level segmentation masks, but we also release coordinates of polygons enclosing individual fashion items. 
The polygons record the shape information of object boundary and serve as an alternative way to segment objects. 
Rectangular bounding box annotations are inferred from polygons to enable fashion item detection.

With fine-grained pixel-level masks, polygons and bounding boxes, we conduct various experiments to demonstrate the usefulness of the ModaNet dataset. Specifically, we investigate and compare the performance of several state-of-the-art deep neural networks for detection and semantic segmentation on this dataset, along with in-depth analysis and discussions. 
Additionally, we train a network to perform polygon prediction and verify its capability of capturing the boundary of objects, posing a potential research direction for fashion item understanding. 
From extensive experimental validation, we show that the proposed dataset enables multiple tasks to achieve promising results, which are difficult to obtain from small-scale and weakly annotated datasets.

\begin{table*}[t]
\centering
\caption{Comparison of ModaNet with other datasets for fashion parsing. ModaNet surpasses previous datasets in terms of annotation granularity and scale. \checkmark$^*$  indicates the annotations are not included in the original dataset. The count of categories excludes non-fashion categories, such as \emph{hair}, \emph{skin}, \emph{face}, \emph{background} and \emph{null}.}
\label{tab:datasets}
\begin{tabular}{@{}lcccccc@{}}
\hline
                  & DeepFashion~\cite{DBLP:conf/cvpr/LiuLQWT16} & CFPD~\cite{DBLP:journals/tmm/LiuFDXHHY14}    & CCP~\cite{DBLP:conf/cvpr/YangLL14}     & Fashionista~\cite{DBLP:conf/cvpr/YamaguchiKOB12} & HPW\cite{DBLP:journals/pami/Liang0SYLDLY15} & ModaNet  \\
\hline
\# of images      & $800,000$   & $2,682$ & $1,004$ & $685$       & $1,833$ & $55,176$ \\
\# of categories  &     50        &     19    &      56   &       53      &  11  & 13  \\
Pixel annotation & \texttimes      & \checkmark    &\checkmark & \checkmark      &\checkmark & \checkmark      \\
Bounding box      & landmarks   &  \checkmark$^*$       & \checkmark$^*$         &     \checkmark$^*$         & \checkmark$^*$  & \checkmark      \\
Polygon           & \texttimes        & \texttimes     & \texttimes     & \texttimes        &\texttimes  & \checkmark      \\ 
\hline
\end{tabular}
\end{table*}

\section{Related Work}

\subsection{Datasets}
Numerous datasets specifically tailored for fashion have been introduced and contributed to the advancement of various vision algorithms. 
Some datasets contain only a particular fashion category, such as the UT Zappos50K dataset~\cite{DBLP:conf/cvpr/YuG14}, while some include more types of fashion items~\cite{DBLP:conf/iccv/HuangFCY15,DBLP:conf/iccv/KiapourHLBB15}. Complementary information like attributes and landmarks are also provided in some datasets for more fine-grained analysis. 
These datasets are mostly designed for image retrieval where the task is to match similar clothes, thus often consist of a large number of images from different domains such as street fashion and product photos. For example, a recent DeepFashion dataset contains as many as $800,000$ images~\cite{DBLP:conf/cvpr/LiuLQWT16}.

On the other hand, some datasets aim at parsing individual fashion items given a street photo image~\cite{DBLP:conf/cvpr/YamaguchiKOB12,DBLP:conf/iccv/YamaguchiKB13,DBLP:journals/tmm/LiuFDXHHY14,DBLP:conf/cvpr/YangLL14,DBLP:journals/pami/Liang0SYLDLY15}. Different from the datasets used for image retrieval that only have image-level labels, these datasets have pixel-level annotations for each type of category, which can be used as segmentation masks to train segmentation models. Although annotations are richer, these datasets usually contain only hundreds or thousands of images due to the difficulty of acquiring pixel-level annotations. 
Therefore, although they are useful for identifying fashion items more accurately, they are not especially helpful for real-world applications, where the diversity of fashion items is enormous regarding type, appearance, composition, and style, \etc.

In contrast to the afore-mentioned datasets, our ModaNet dataset advances the quality of data regarding both scale and granularity of annotations. We have more than $55,000$ fully-annotated images with pixel-level segments, polygons and bounding boxes covering 13 categories. 
A comparison is shown in Table~\ref{tab:datasets}.

\subsection{Detection}
There are various applications based on existing datasets that try to understanding fashion from different perspectives. 
One important task is to detect fashion items from images.
Recently, various approaches based on deep neural networks have been proposed for generic object detection and achieved promising results, among which, some representative works are RCNN~\cite{DBLP:conf/cvpr/GirshickDDM14}, Fast RCNN~\cite{DBLP:conf/iccv/Girshick15}, Faster RCNN~\cite{Ren/fasterrcnn2015}, SSD~\cite{Liu/ssdeccv2016}, R-FCN~\cite{Dai/cvpr2016}, YOLO~\cite{Redmon/cvpr2016}, \emph{etc}. 
To improve the performance of these detectors, several modifications have been proposed. Shrivastava \etal~\cite{DBLP:conf/cvpr/ShrivastavaGG16} propose to use online hard negative mining to adaptively select diverse, high-loss samples for training. 
Lin \etal~\cite{DBLP:conf/iccv/LinGGHD17} propose the focal loss as an alternative way to do hard negative mining to alleviate the effect of overwhelming negative samples.
Bodla \etal~\cite{DBLP:conf/iccv/BodlaSCD17} introduce a soft-NMS to replace the traditional non-maximal suppression (NMS) used in object detection to discount the confidence score of predicted boxes rather than completely discarding them.  
By changing the backbone networks used in the detector, feature pyramid network~\cite{DBLP:conf/cvpr/LinDGHHB17}, also called RetinaNet, has improved the detection accuracy, where feature maps from different convolutional layers are fused to provide more discriminative power. 
Deformable convolutional networks~\cite{DBLP:conf/iccv/DaiQXLZHW17} also greatly improves detection performance for non-rigid objects by allowing convolution to operate on irregular regions instead of a grid.
Regarding fashion item detection specifically, Hara \etal~\cite{DBLP:conf/wacv/HaraJP16} incorporate contextual information from body poses to guide detection and extract features from an off-the-shelf deep neural network. 
Liu \etal~\cite{DBLP:conf/cvpr/LiuLQWT16} present preliminary results using Fast RCNN trained on the DeepFashion dataset, but the model can only detect upper-body, 
lower-body and full-body objects due to the lack of fine-grained annotations.

\subsection{Segmentation}
The main focus of semantic image segmentation is to assign an object label to each pixel in an image.  
Recently, researchers have also begun tackling new problems such as instance segmentation~\cite{Hariharan/eccv2014}, which aims to assign a unique identifier and a semantic meaningful label to each segmented object in the image. Semantic segmentation has traditionally been approached using probabilistic models known as Conditional Random Fields (CRFs)~\cite{He/cvpr2014}, which explicitly model the correlations among the pixels being predicted. However, in the recent years, learning a better feature representation~\cite{Shotton/eccv2006,Ladicky/cvpr2009} has shown to play an important role in pushing the state-of-the-art performance of semantic image segmentation. The rise of deep Convolutional Neural Networks has significant improved the way of learning feature representation. In particularly, the fully convolutional networks (FCNs) have shown significantly performance boost in semantic image segmentation~\cite{Shelhamer/pami2016}. There are two directions of improving semantic image segmentation.
One is to improve the architectures or the bottleneck module of neural networks. The representative work is Chen~\etal~\cite{Chen/pami2018}, which has further developed the FCNs using atrous layers. It has also used densely-connected CRFs as a post-processing step. 
Yu~\etal~\cite{Yu/iclr2016} has improved semantic image segmentation by introducing dilated convolution, which increases the resolution of output feature maps without reducing the receptive field of individual neurons. 
Zhao~\cite{Zhao/cvpr2017} has proposed a pyramid scene parsing network that explores the prior global representation to produce good results on scene parsing task. 
Chen~\cite{Chen/cvpr2018} has developed modules that employ the atrous convolution in cascade or in parallel to capture multi-scale context by adopting multiple atrous rates, which has shown state-of-the-art performance in PASCAL VOC benchmark. The other direction is to incorporate CRFs into an end-to-end trainable framework for semantic image segmentation, with the hope that joint training would help improve performance further. 
Zheng~\etal~\cite{Zheng/iccv2015} formulate an end-to-end trainable framework using the fully convolutional neural networks and densely-connected CRFs, while Liu~\etal~\cite{Liu/pami2017}, Schwing~\etal~\cite{Schwing/arxiv2015} and Lin~\etal~\cite{Lin/cvpr2017} have explored similar ideas along this line. 
These approaches are developed for generic object categories, and some are developed for scene parsing. Different from generic objects or scenes, there are more self-occlusion in fashion images.
One goal of the ModaNet is to facilitate future semantic image segmentation that would perform well in fashion. 

Several works have also explored semantic image segmentation in the context of fashion. 
Yamaguchi~\etal~\cite{DBLP:conf/cvpr/YamaguchiKOB12} has pioneered the work in clothing parsing, and later further improved the fashion parsing performance by using retrieval-based approach~\cite{DBLP:conf/iccv/YamaguchiKB13}. Dong~\etal~\cite{Dong/iccv2013} use Parselets as the building blocks of the parsing model. 
Liu~\etal~\cite{Liu/acmm2015} present a solution that harnesses the context in fashion videos to boost the fashion parsing performance. Liang~\etal~\cite{Liang/iccv2015} developed a convolutional neural network approach for human parsing. These works can be categorized into human parsing and clothing parsing. 
Clothing parsing attempts to identify the fine-grained categories of clothing items such as t-shirt and blouse, whereas human parsing tries to identify the body parts and broad clothing categories. The ModaNet dataset focuses on the clothing fashion parsing rather than human parsing.

\section{The ModaNet Dataset}

\subsection{Constructing ModaNet}

We first collect 1 million images from the PaperDoll dataset~\cite{DBLP:conf/iccv/YamaguchiKB13}, which are not fully annotated for object detection and semantic image segmentation purpose. 
This dataset has a large variety of images that are relevant to many fashion applications including street-to-shop and the shop-the-look.

Second, we apply Faster RCNN~\cite{Ren/fasterrcnn2015} that is pretrained on COCO dataset~\cite{Tsungyi/eccv2014} to detect if there is only one person present in the image, and only collect those images with a single person. 
While addressing the images containing multiple persons is significantly more challenging due to occlusion and scale variance, we leave it in our future work and focus on dealing with the images containing only one person in the first release.

Among the initial set of selected images, we further manually select $2,000$ images that are not suitable for annotations due to low image quality and $2,000$ images that are high quality for annotation. 
On these $4,000$ images, we fine-tune the last layer of a ResNet-50~\cite{Kaiming/cvpr2016} model pre-trained on ImageNet~\cite{Deng/cvpr2009} as a classifier for image quality.
We apply this classifier to the entire set of initially selected images to classify them into different levels of quality, and select those images of high quality and containing only one person. 
This step is necessary to reduce the number of images of the bad quality that are later sent to human annotators.

Finally, we send these images to human annotators. 
The tasks for human annotators are two-fold: one is to skip images they think is ambiguous to annotate, and the other one is to provide polygon annotations for individual objects in the image and assign a label from a pre-defined set of fashion object categories. We annotated this dataset through a collaboration with data annotation organization. We have 17 annotators (9 female and 8 male) who were trained for 2 weeks before starting annotating. After training, all the annotators reached 99.5\% internal quality accuracy, which was manually rated by 2 supervisors with computer vision background. The project was monitored daily and also verified weekly.
The annotation system was designed to allow multiple annotators to cross check the same annotations. The annotators delivered polygon annotations based on the image, while their supervisor manually checked if the image was annotated correctly.
During the training phase, some annotators annotated images that were too blurry or dark, while other annotators annotated images containing no person or only the feet. We have fixed these issues by providing various sample images that they should skip or annotate. With polygon annotations, we can generate the ground truth for multiple computer vision tasks including object detection and semantic image segmentation.


\subsection{Statistics}

\begin{figure*}[t]
 \centering
 \setlength{\tabcolsep}{0.1pt}
 \setlength{\fboxsep}{0pt}%
 \setlength{\fboxrule}{0.1pt}%
 \renewcommand{\arraystretch}{0.6}
 \begin{tabular}{cccccccccc}
 \includegraphics[width=.1\textwidth]{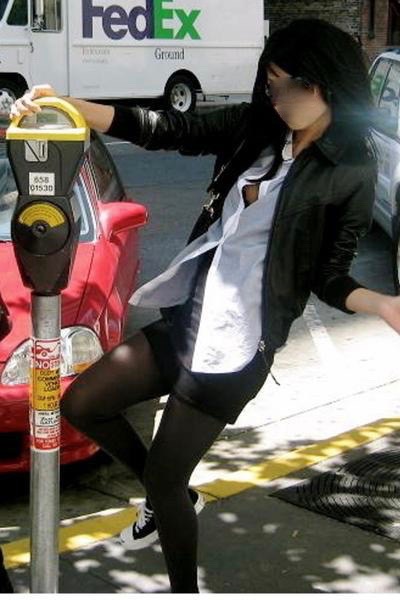} & 
 \includegraphics[width=.1\textwidth]{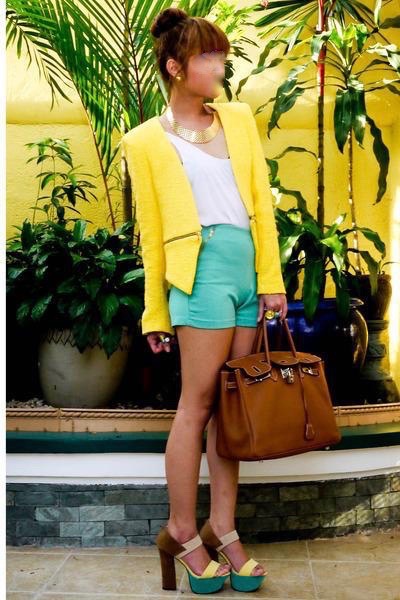} &
 \includegraphics[width=.1\textwidth]{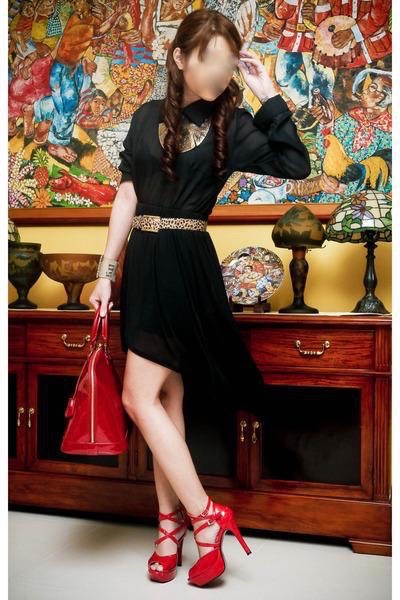} &
 \includegraphics[width=.1\textwidth]{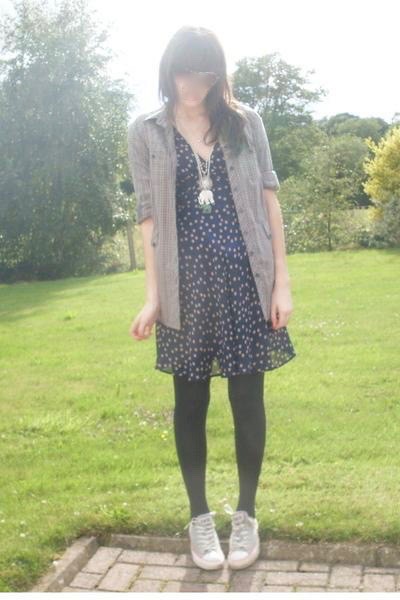} &
 \includegraphics[width=.1\textwidth]{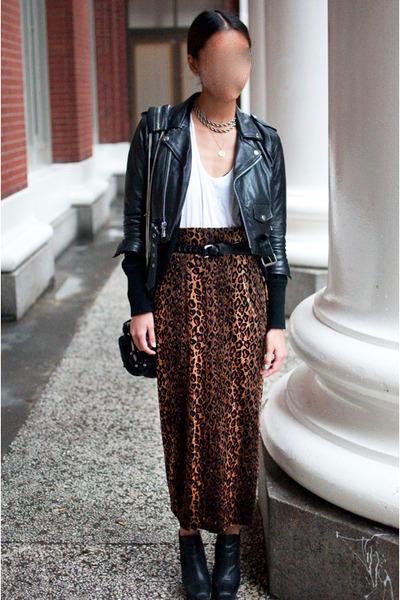} & 
 \includegraphics[width=.1\textwidth]{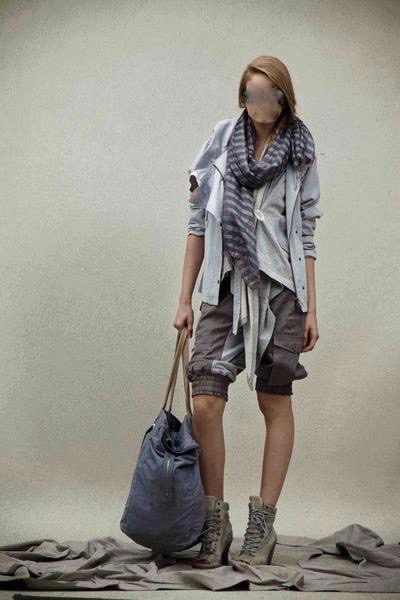} &
 \includegraphics[width=.1\textwidth]{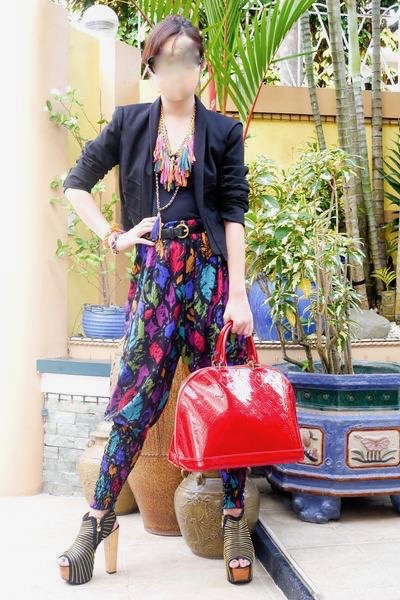} &
 \includegraphics[width=.1\textwidth]{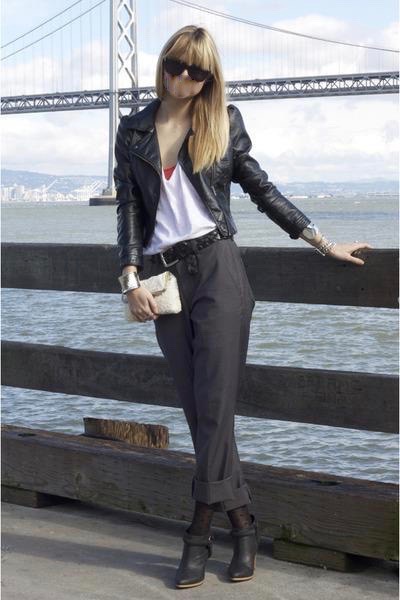} &
 \includegraphics[width=.1\textwidth]{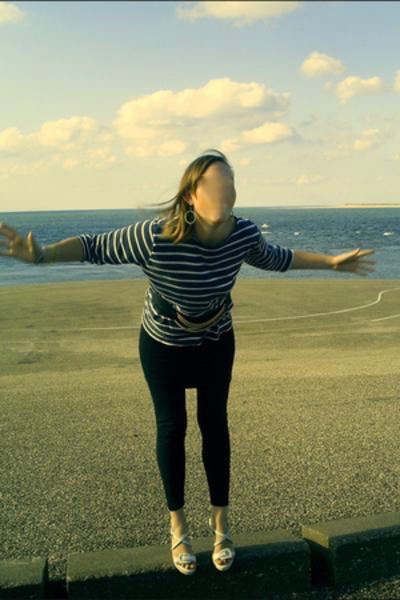} &
 \includegraphics[width=.1\textwidth]{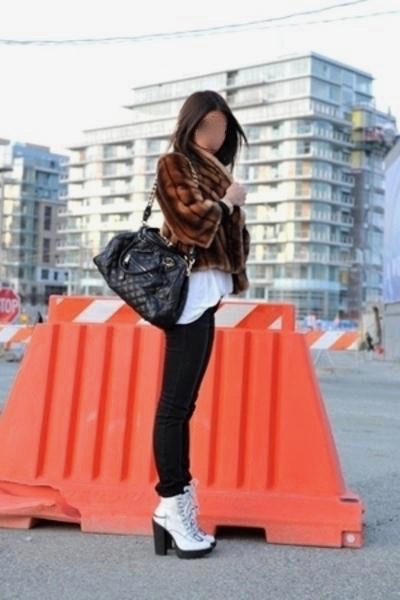}\\
 \includegraphics[width=.1\textwidth]{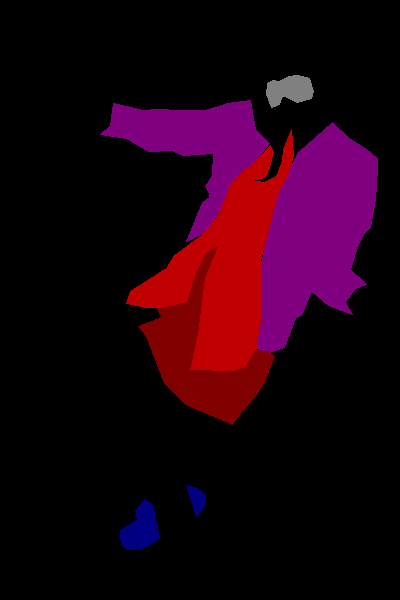} & 
 \includegraphics[width=.1\textwidth]{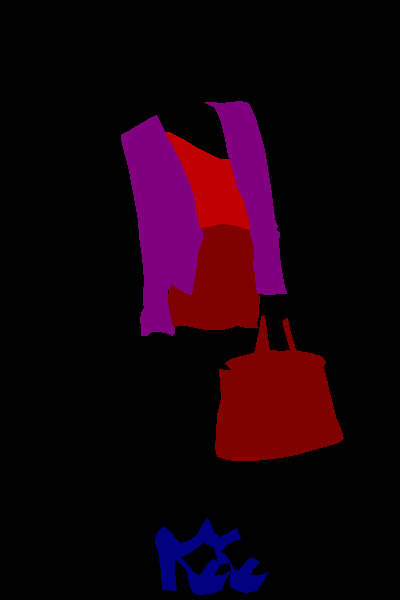} &
 \includegraphics[width=.1\textwidth]{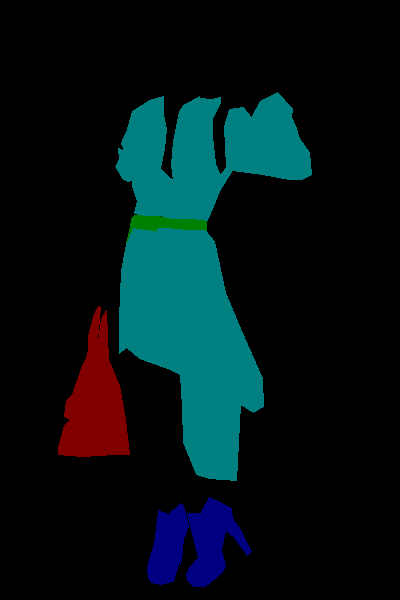} &
 \includegraphics[width=.1\textwidth]{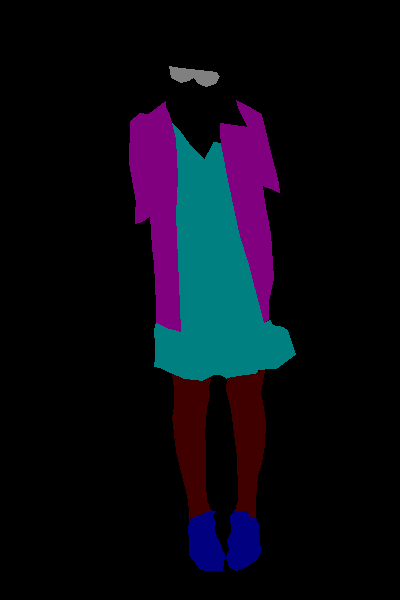} &
 \includegraphics[width=.1\textwidth]{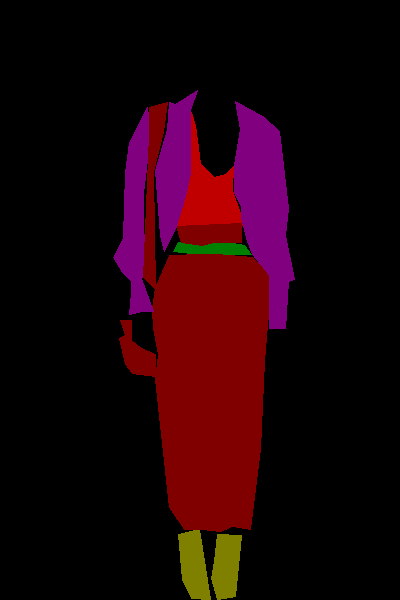} & 
 \includegraphics[width=.1\textwidth]{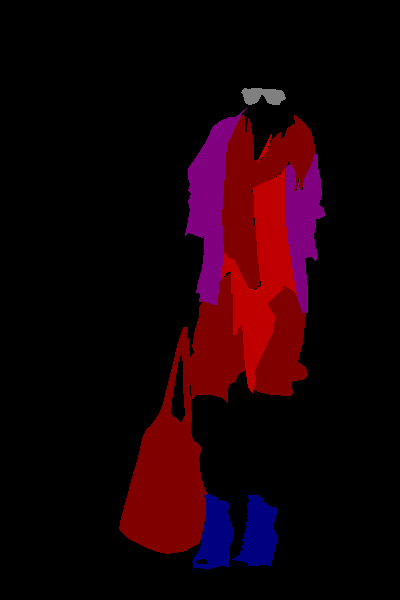} &
 \includegraphics[width=.1\textwidth]{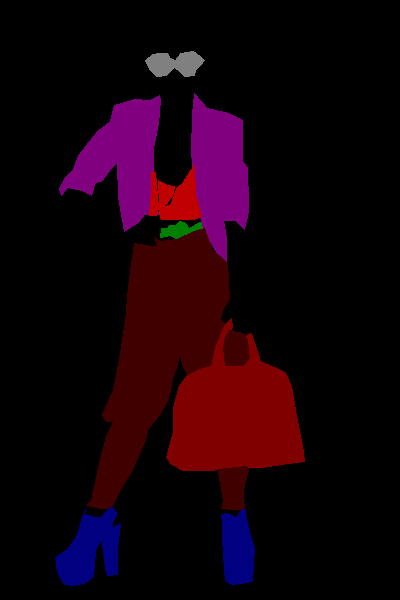} &
 \includegraphics[width=.1\textwidth]{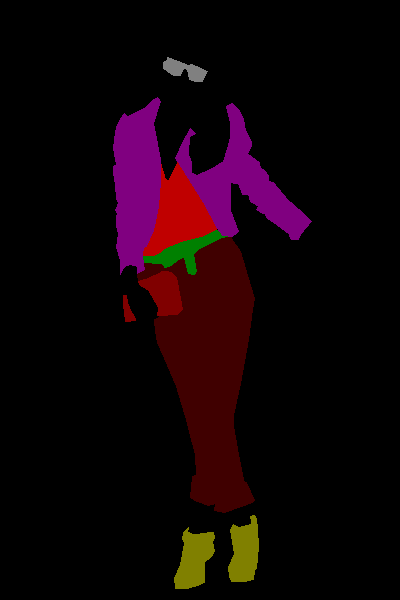} &
 \includegraphics[width=.1\textwidth]{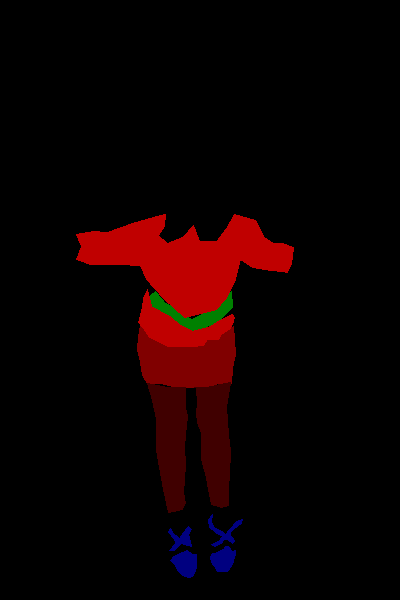} &
 \includegraphics[width=.1\textwidth]{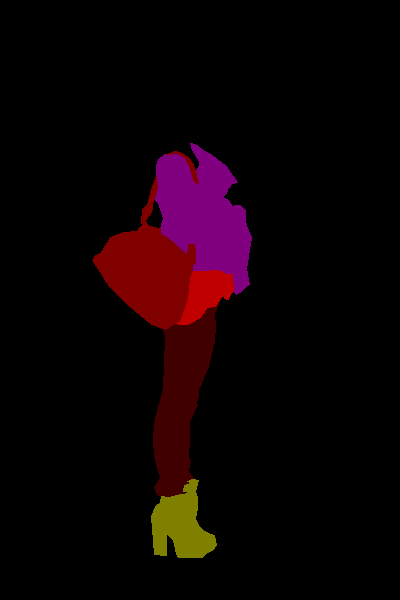}\\
 \includegraphics[width=.1\textwidth]{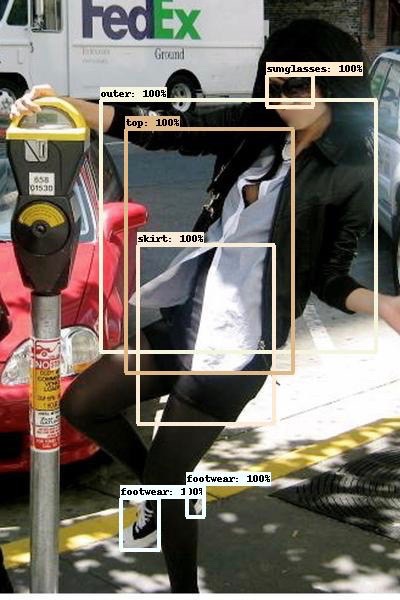} & 
 \includegraphics[width=.1\textwidth]{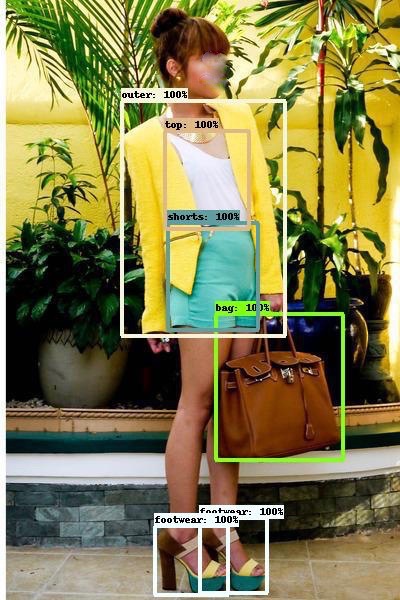} &
 \includegraphics[width=.1\textwidth]{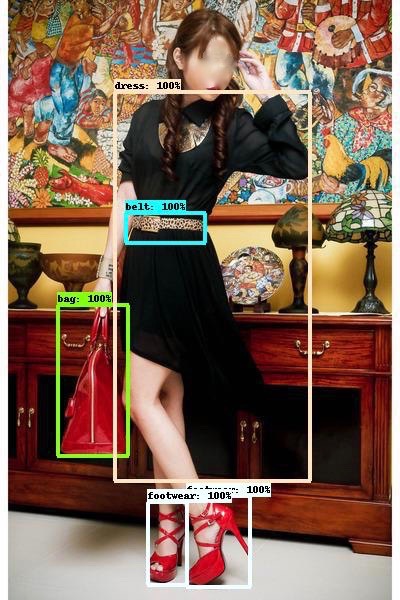} &
 \includegraphics[width=.1\textwidth]{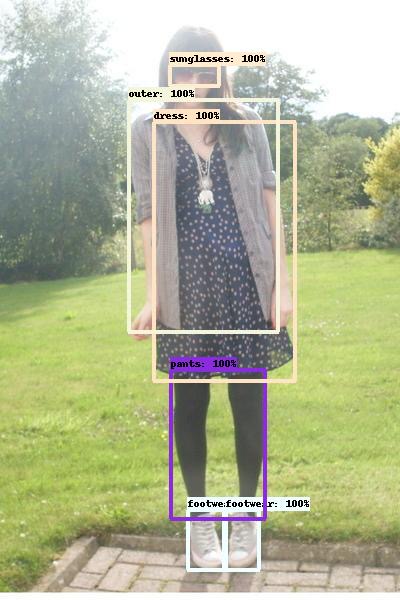} &
 \includegraphics[width=.1\textwidth]{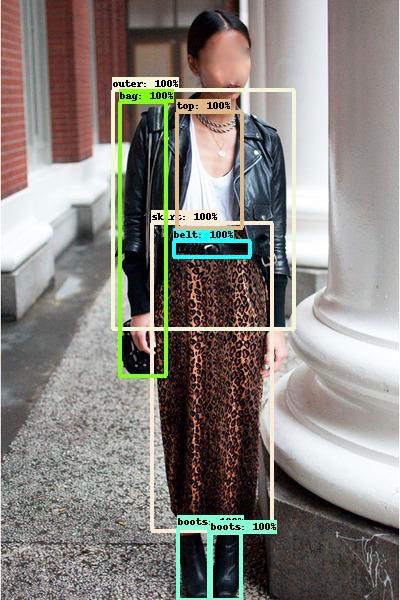} & 
 \includegraphics[width=.1\textwidth]{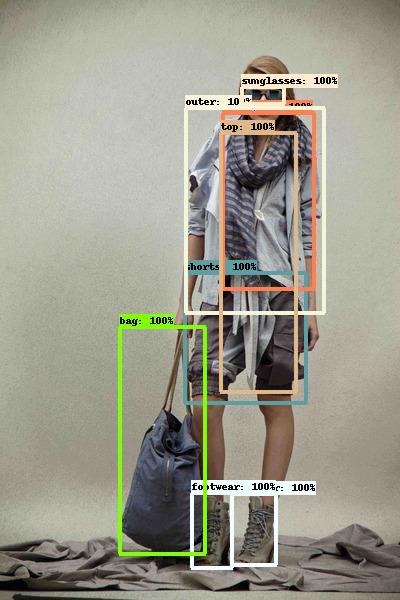} &
 \includegraphics[width=.1\textwidth]{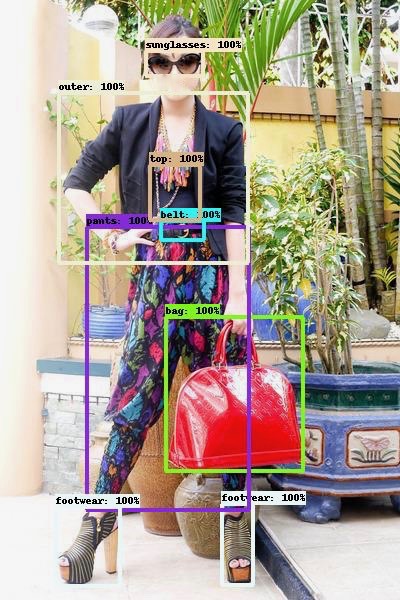} &
 \includegraphics[width=.1\textwidth]{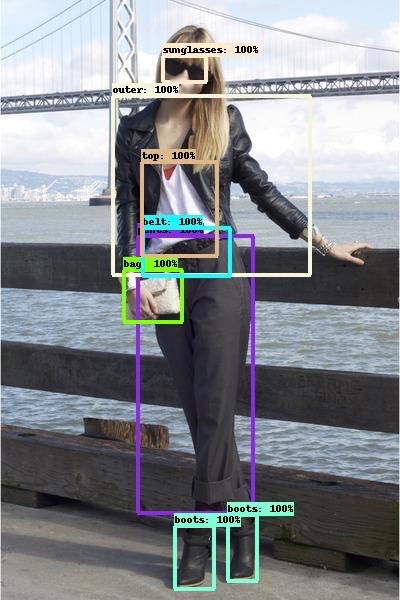} &
 \includegraphics[width=.1\textwidth]{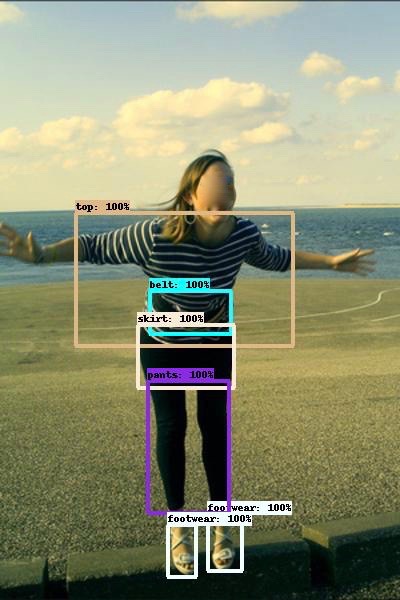} &
 \includegraphics[width=.1\textwidth]{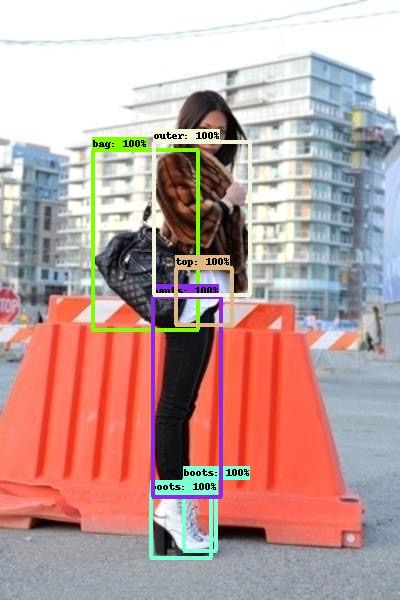}\\
 
 \end{tabular}
 \caption{Examples of original images with corresponding pixel-level segmentation masks and bounding box annotations of the proposed ModaNet dataset. The first row shows the original color street image containing a person with fashion products. The second row indicates the pixel-wise annotation for the fashion product, where color is encoded to represent the fashion product category. The third row represents the bounding box annotation overlaying on the color images. Best view in color.}
\label{fig:annotations} 
\end{figure*}

\begin{figure*}[tpb]
\begin{center}
\includegraphics[width=.32\linewidth]{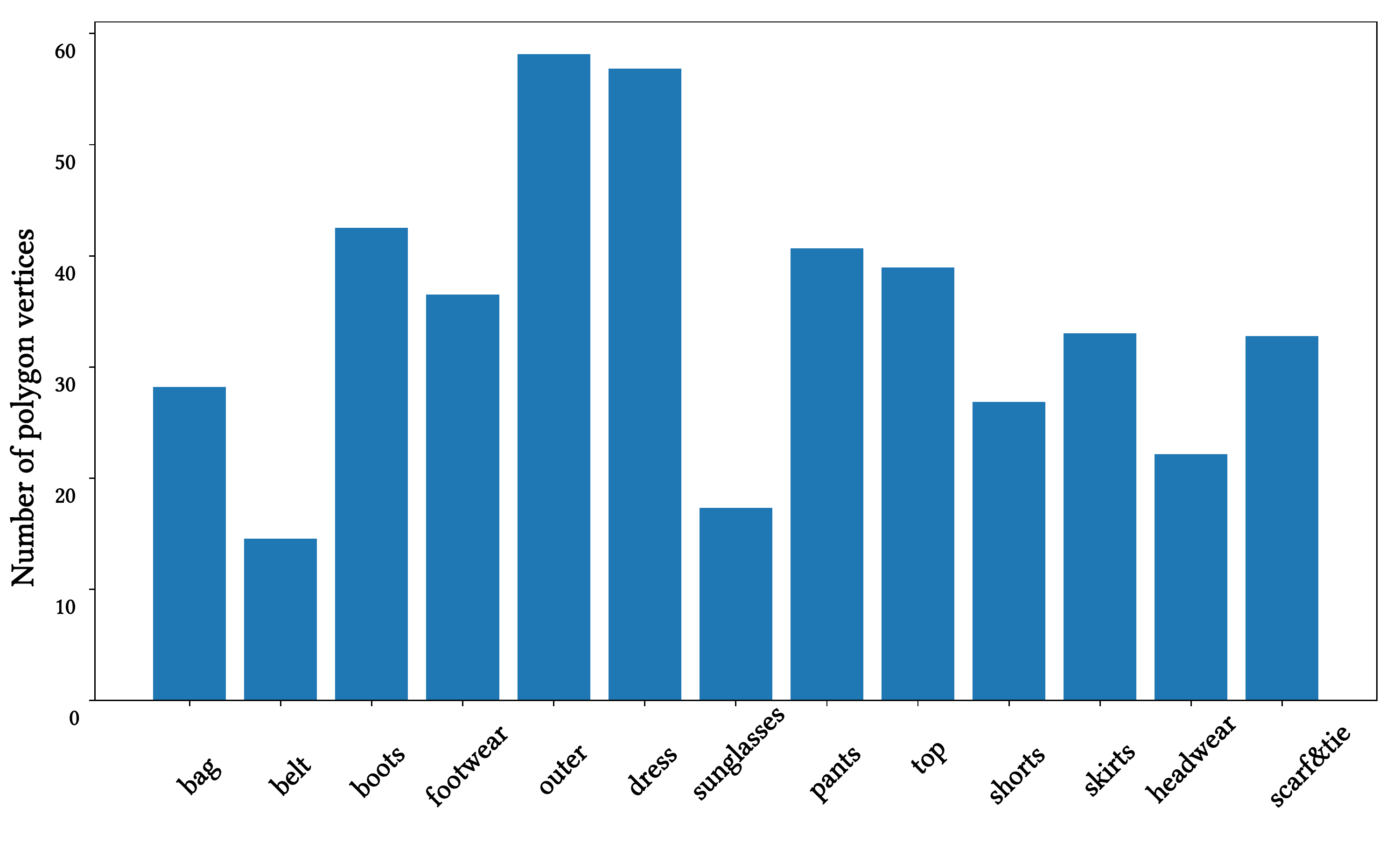}
\includegraphics[width=.32\linewidth]{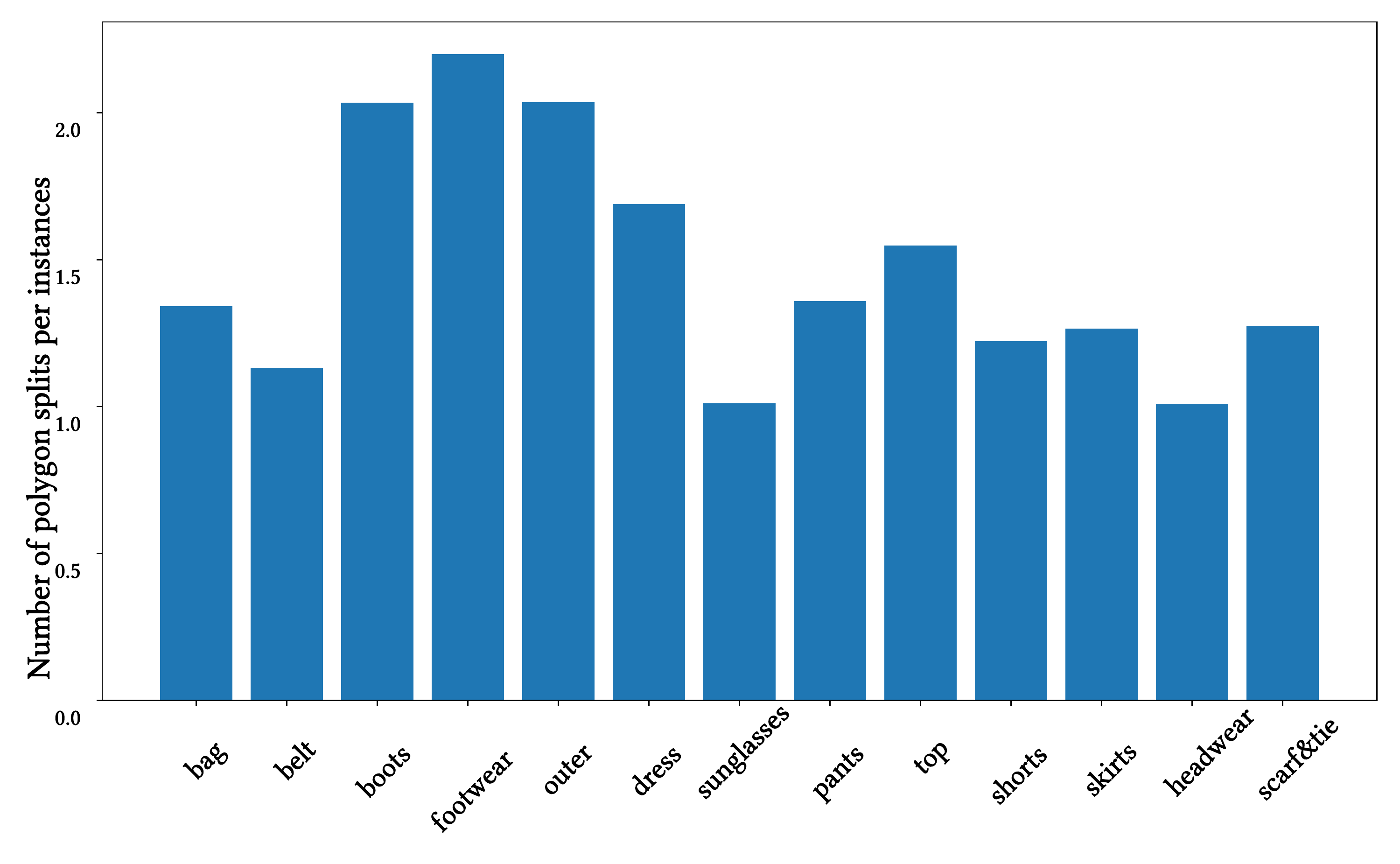}
\includegraphics[width=.32\linewidth]{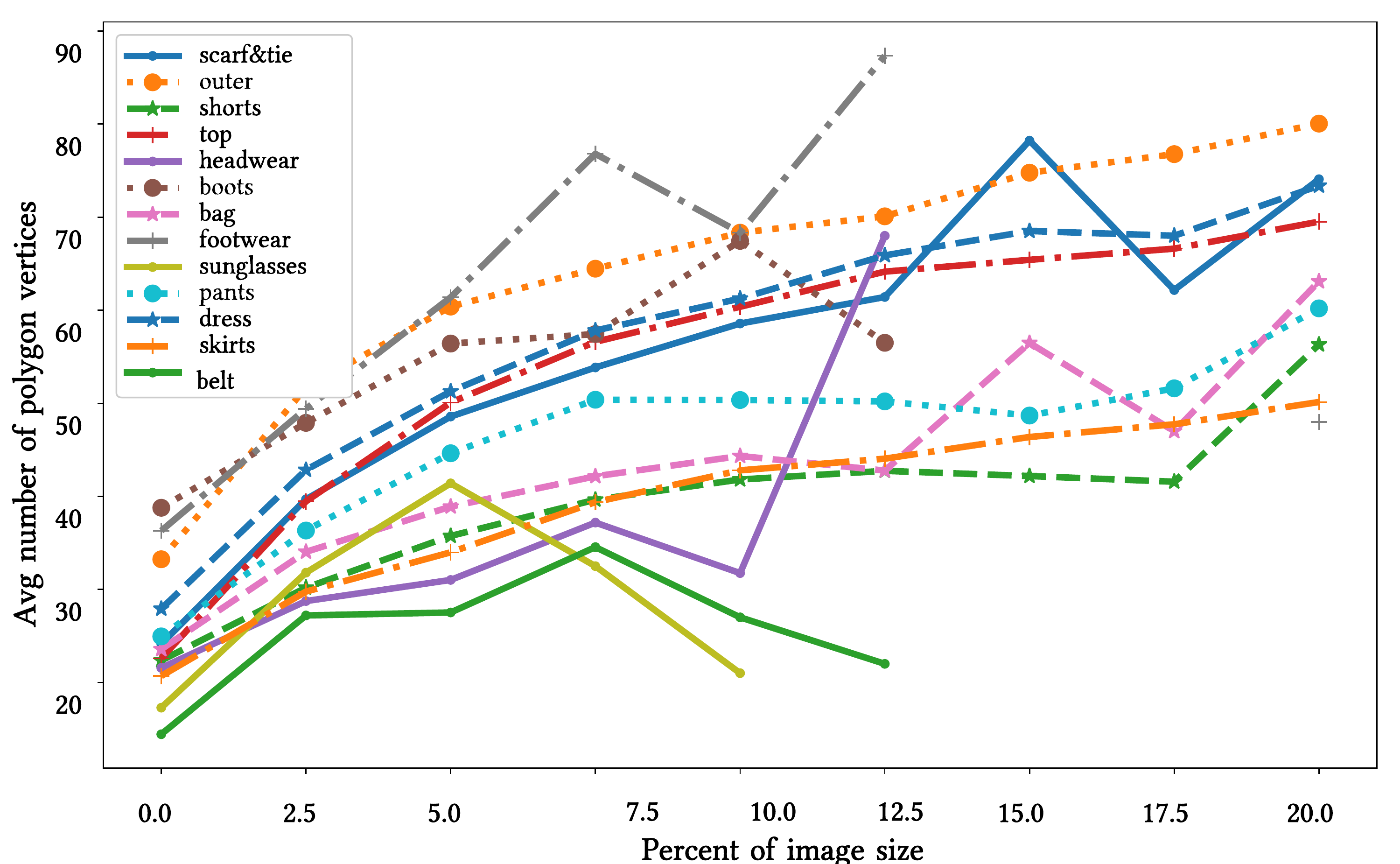}
\end{center}
\vspace{-2mm}
\caption{Annotation polygon statistics. Left: average number of vertices in polygon annotations for each category. Middle: average number of segments in polygon annotations of an instance of each category. Right: total number of vertices in the annotation polygons for each instance vs the size of the instance. Best viewed in color.}
\label{fig:polygon_distribution}
\end{figure*}

The dataset contains $13$ meta categories, where each meta category groups highly related categories to reduce the ambiguity in the annotation process. 
The $13$ meta categories are \emph{bag}, \emph{belt}, \emph{boots}, \emph{footwear}, \emph{outer}, \emph{dress}, \emph{sunglasses}, \emph{pants}, \emph{top}, \emph{shorts}, \emph{skirt}, \emph{headwear}, \emph{scarf\&tie}. 
The detailed mapping from the original labels to the meta categories is shown in Table~\ref{tab:modanet statistics}. 
Some examples of original images, pixel-level segmentation masks, and bounding boxes are shown in Figure~\ref{fig:annotations}. We also visualize the distribution of the clothing categories in this dataset in Figure~\ref{tab:modanet statistics}. The most common objects in this dataset are \emph{footwear}, \emph{top}, \emph{outer}, \emph{pants}, and \emph{bags}. Most images contain 3 to 5 fashion objects. We also see that categories such as \emph{belts}, \emph{headwear}, and \emph{scarf\&tie} account for a small portion of the outfit, while \emph{dress}, \emph{outer}, \emph{pants} and \emph{skirts} fall on the large side of the instance size spectrum.
Since the ModaNet dataset was annotated using polygons, we present the statistics of polygon annotations in Figure~\ref{fig:polygon_distribution}. 
We can see that categories such as \emph{outer}, \emph{dress}, \emph{boots} and \emph{pants} require the most number of clicks to annotate due to their size and shape, while \emph{sunglasses} and \emph{belts} have the least number of vertices in their polygons. 
Additionally, a clothing instance can have a lot of details, \eg, straps on heels or it can be occluded by body parts or other clothing layers, hence requiring multiple polygons to annotate it fully. We see that \emph{footwear}, \emph{outer} and \emph{boots} require the most number of polygon segments, while \emph{sunglasses} and \emph{belts} are captured in one polygon. Finally, we observe that with the size of instances growing, the number of vertices increases correspondingly, with \emph{belts} and \emph{sunglasses} as exceptions hinting at their less complex shape characteristics.

\begin{table}
\centering
\small
\caption{\textbf{ModaNet statistics}. We group highly-related categories to form 13 meta categories.}
\label{tab:modanet statistics}
\begin{tabularx}{\linewidth}{@{}lXccc@{}}
\hline
Meta & Raw & \#Train & \#Val & Avg Inst. size\\
\hline
bag  & bag & $36,699$  & $2,155$  & 4.88\% \\
belt  & belt & $13,743$ & $771$ & 0.46\% \\
boots  & boots & $7,068$ & $691$ & 2.40\%  \\
footwear  & footwear & $39,364$ & $1,617$ & 0.96\% \\
outer  & coat, jacket, suit, blazers
& $23,743$ & $1,358$ & 7.48\% \\
dress  & dress, t-shirt dress & $14,460$ & $804$ & 10.49\% \\
sunglasses  & sunglasses &  $8,780$ & $524$ & 0.31\% \\
pants  & pants, jeans, leggings & $23,075$  &  $1,172$ & 5.65\% \\
top  & top, blouse, t-shirt, shirt & $34,745$ & $1,862$ & 4.83\% \\ 
shorts  & shorts & $5,775$ & $429$ & 2.86\% \\ 
skirt  & skirt & $10,860$  & $555$ &  6.40\% \\ 
headwear  & headwear & $5,405$ & $491$ & 1.25\% \\ 
scarf\&tie  & scarf, tie & $3,990$ & $378$ & 2.55\%  \\ 
\hline
\end{tabularx}
\end{table}

\begin{figure*}[t]
\begin{center}
\includegraphics[width=.32\linewidth]{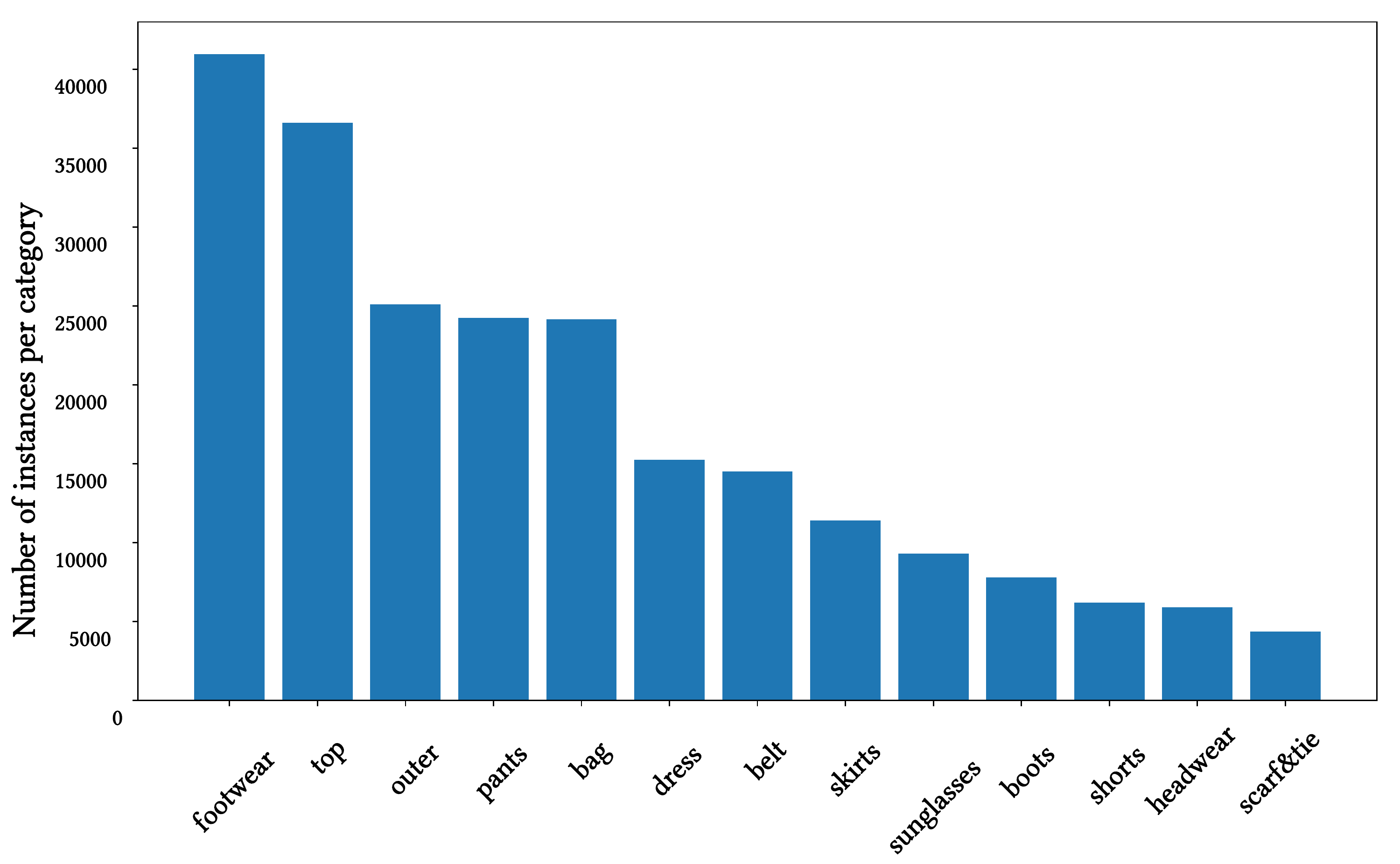}
\includegraphics[width=.32\linewidth]{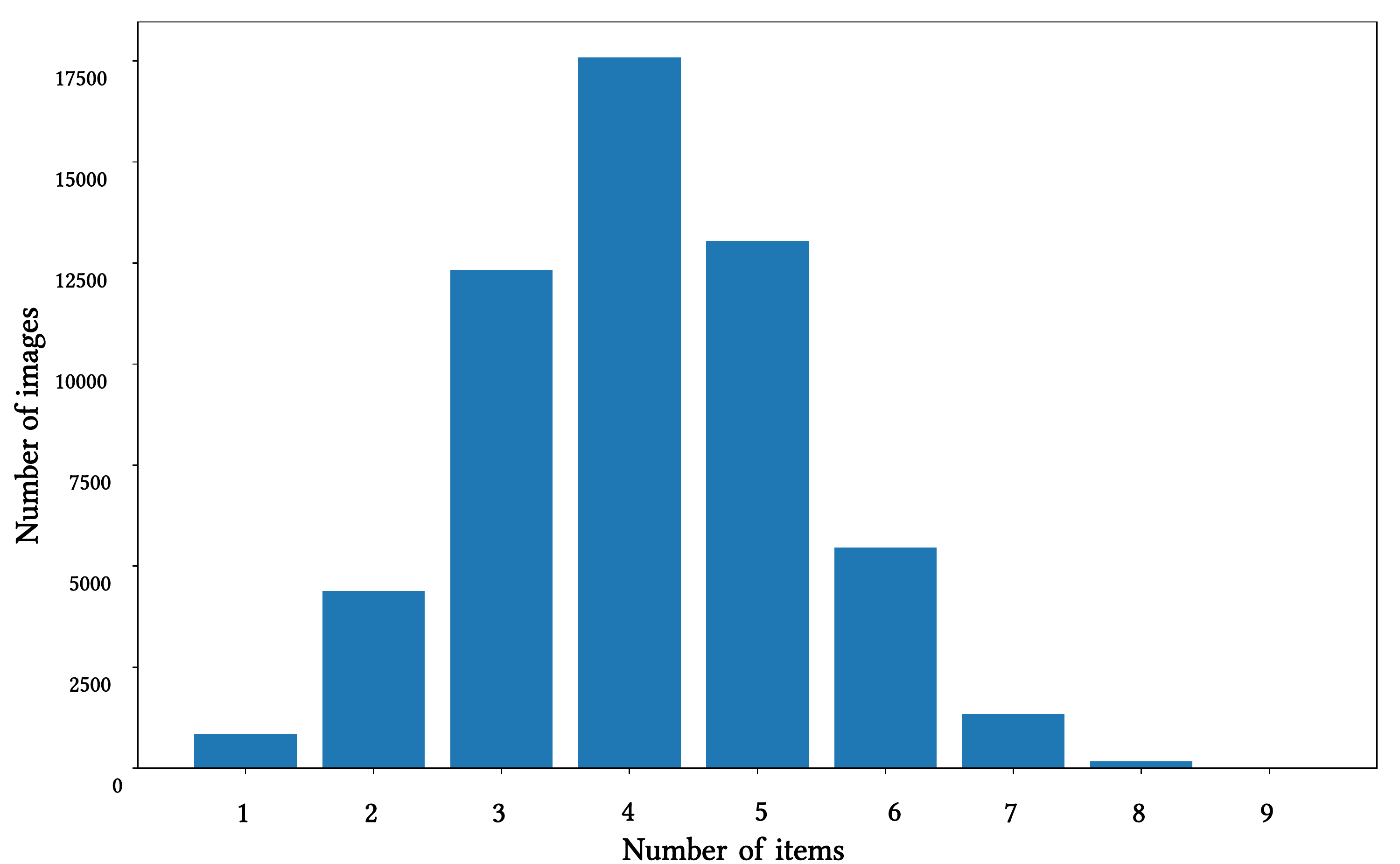}
\includegraphics[width=.32\linewidth]{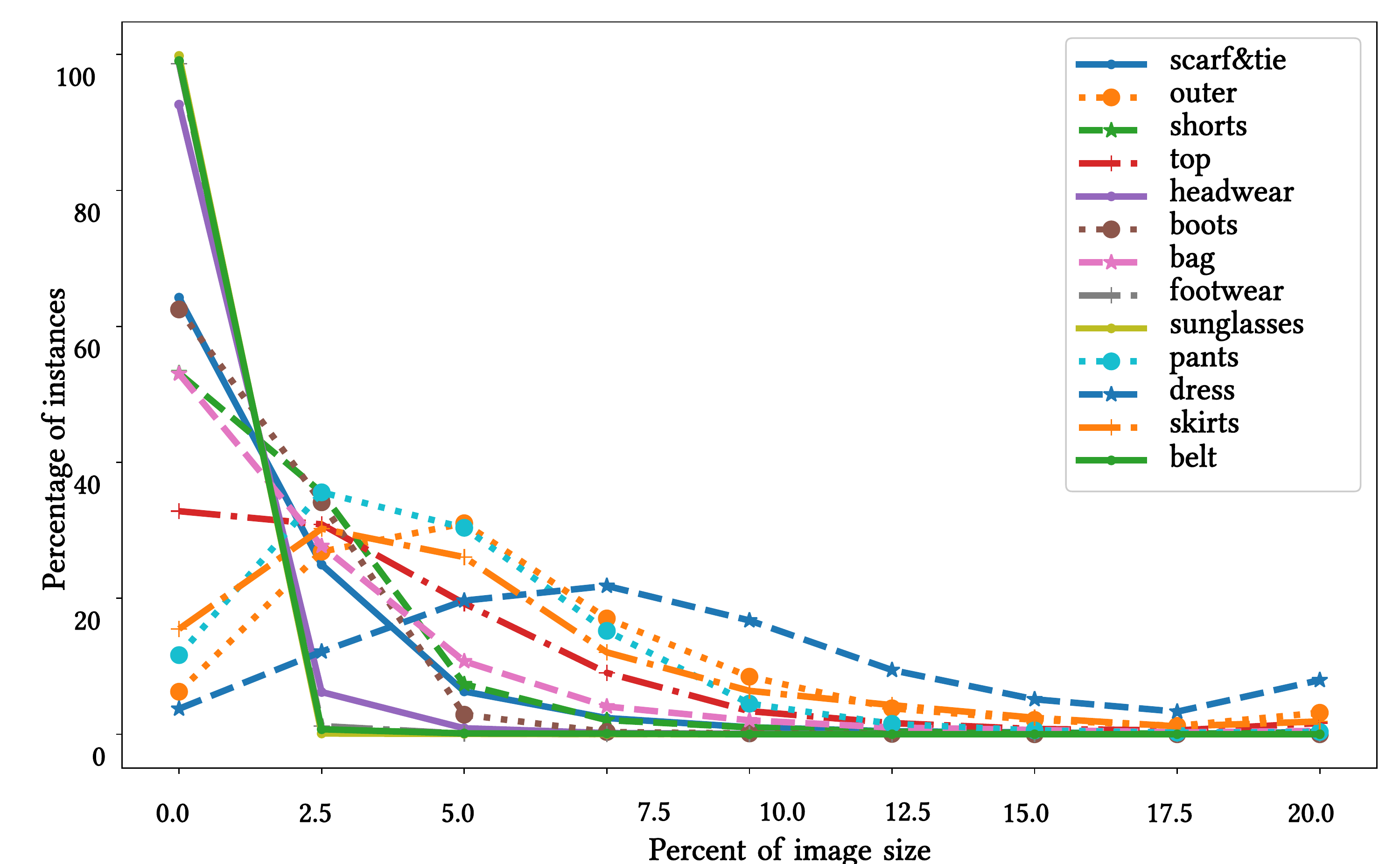}
\end{center}
\vspace{-2mm}
\caption{Dataset statistics. Left: the number of instances per category in descending order. Middle: the distribution of images from minimum number of annotated items per image to the maximum number of items per image. We can see that on average most images contain four annotated items. Right: The distribution of
instance sizes across categories. Best viewed in color.}
\label{fig:dataset_distribution}
\end{figure*}



\section{ModaNet Applications}
\subsection{Object Detection for Fashion}
The dataset enables fashion item detection, where each fashion item is localized and assigned with a category label, which can be further used for visual search and product recommendation. 

\subsubsection{Groundtruth generation}
Given pixel-level and polygon annotations, we can easily infer bounding boxes for training images. In this work, we choose to generate bounding boxes from the polygon annotations as the boundary information is easily accessible. 
While a single fashion item might be annotated by multiple polygons due to composition, pose variation and occlusion in some cases, we assume that there is only one person and each fashion item only appears once (twice for a pair of shoes) and merge them to ensure a single bounding box. 
Also, we discard all boxes with the size smaller than $100$ pixels to ensure the training data is large enough to train a reasonable model. 
We split the whole dataset into training containing $52,377$ images and a validation set containing the remaining $2,799$ images. The split enforces that each category from the validation set contains at least $500$ instances so that the validation accuracy is reliable. 
Note that it is not feasible to enforce that the number of instances in the validation set for each category is precisely proportional to the total number of instances in the dataset due to that multiple items appear in a single image. 
All of our subsequent experiments follow the same training and validation split.
Some example annotations are shown in Figure~\ref{fig:annotations}.
All the bounding box annotations will be made available along with the dataset. 

\subsubsection{Detectors}

While there are numerous object detectors available in the literature, we choose three most popular detectors to evaluate their performance on the dataset, which are Faster RCNN~\cite{Ren/fasterrcnn2015}, SSD~\cite{Liu/ssdeccv2016} and YOLO~\cite{Redmon/cvpr2016}.
Both SSD and YOLO are real-time detectors and have similar network architectures, which can be used for mobile devices. We compare their performance to investigate the trade-off between accuracy and efficiency.
Faster RCNN is more accurate but takes more time during inference, making it more suitable for offline applications.
Specifically, for Faster RCNN, the backbone network is Inception-ResNet~\cite{Szegedy/aaai2017}. We use Inception V2 network~\cite{Szegedy/aaai2017} for SSD and YOLO v2 network for YOLO detector.

\subsubsection{Performance comparison}

We evaluate and compare the performance of Faster RCNN, SSD and YOLO on our ModaNet dataset using mean average precision (mAP) as the evaluation metric. 
Meanwhile, we also show the average precision (AP) per category to understand which categories are more challenging to the detectors. 
Also, we adjust the IOU threshold and compute mAP based on varying IOU thresholds to provide more insights of how well the predicted bounding boxes align with the ground truth as a robustness measurement.

\begin{figure*}[tpb]
\begin{center}
\includegraphics[width=.32\linewidth]{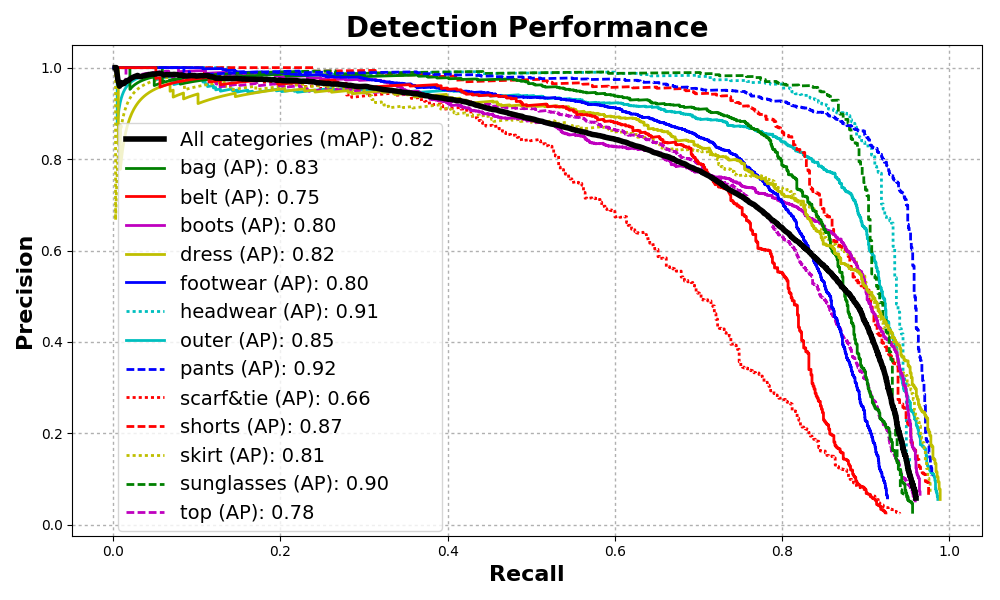}
\includegraphics[width=.32\linewidth]{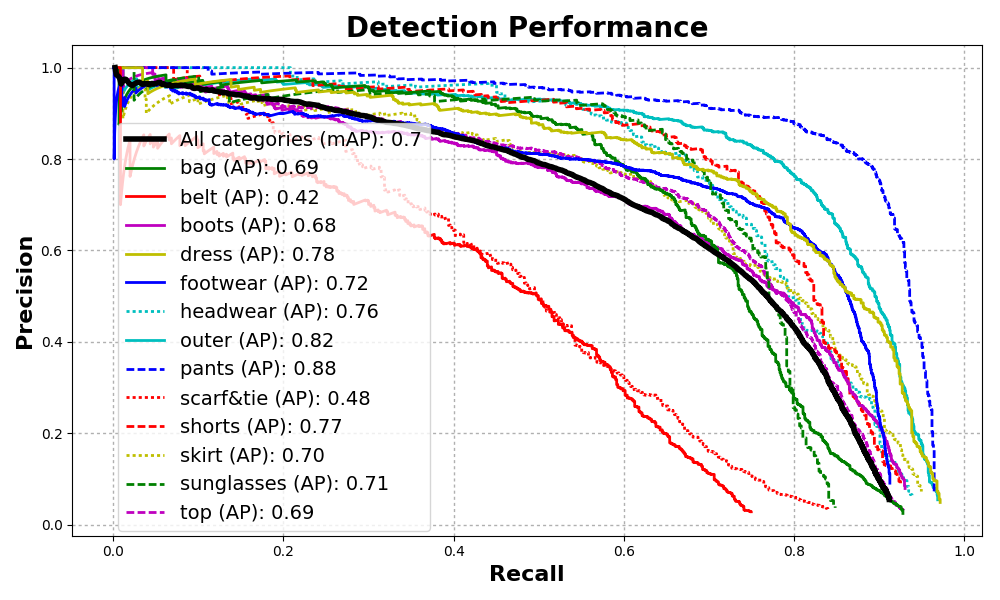}
\includegraphics[width=.32\linewidth]{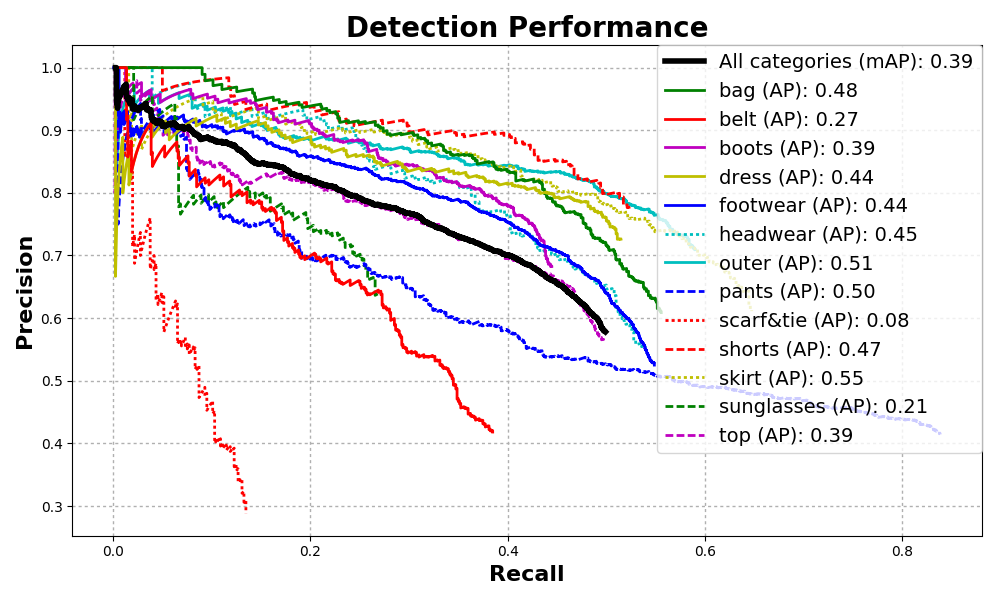}
\end{center}
\vspace{-2mm}
\caption{Performance comparison of Faster RCNN (left), SSD (middle) and YOLO (right). Best view in color.}
\label{fig:map-per-cat}
\end{figure*}

\begin{figure}[t]
 \centering
 \setlength{\tabcolsep}{0.5pt}
 \setlength{\fboxsep}{0pt}%
 \setlength{\fboxrule}{0.1pt}%
 \renewcommand{\arraystretch}{0.6}
 \begin{tabular}{cccc}
GT & Faster RCNN & SSD & YOLO \\
 \includegraphics[width=.11\textwidth]{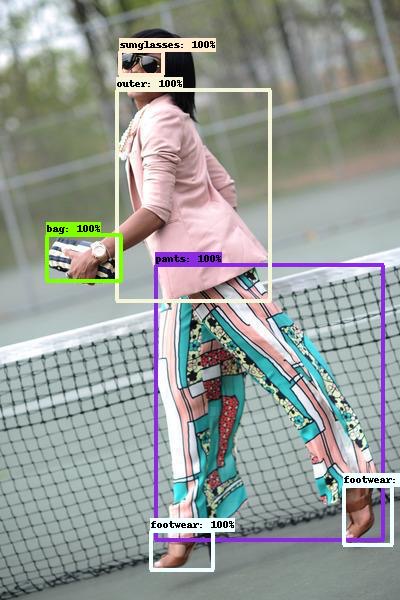} & 
 \includegraphics[width=.11\textwidth]{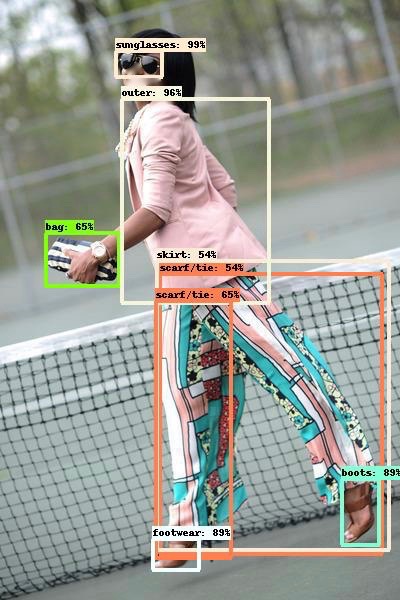} &
 \includegraphics[width=.11\textwidth]{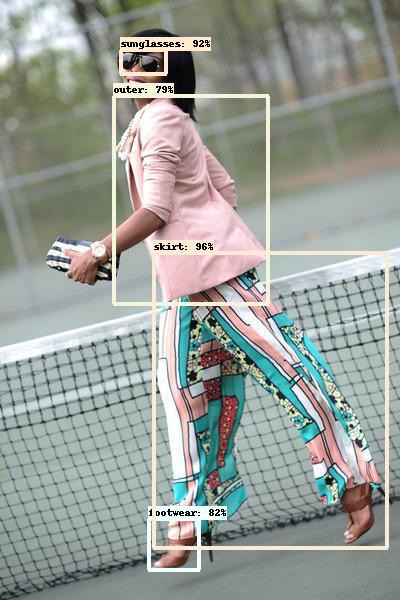} &
 \includegraphics[width=.11\textwidth]{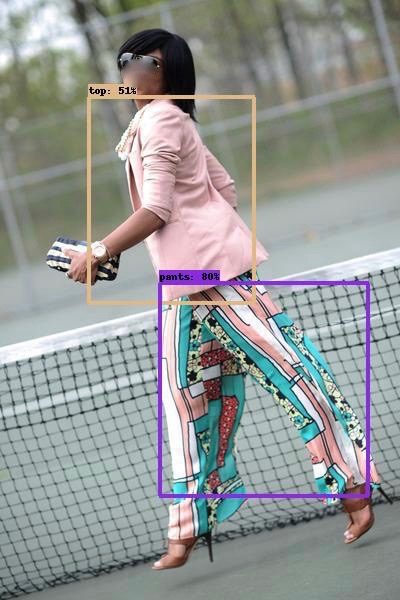}\\

 \includegraphics[width=.11\textwidth]{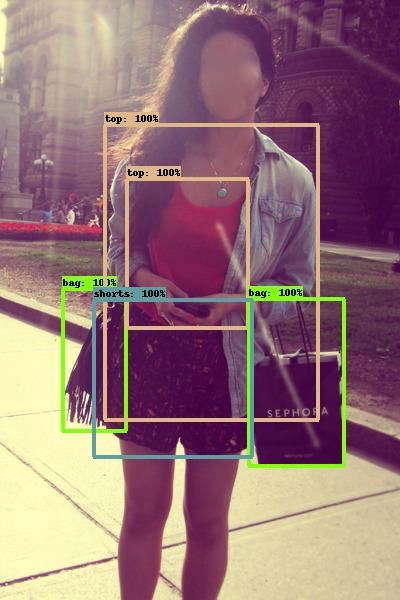} & 
 \includegraphics[width=.11\textwidth]{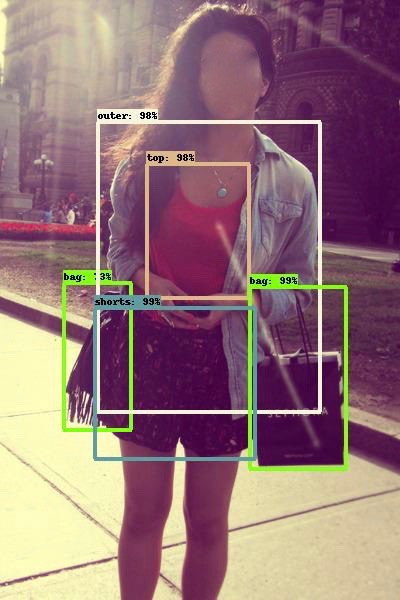} &
 \includegraphics[width=.11\textwidth]{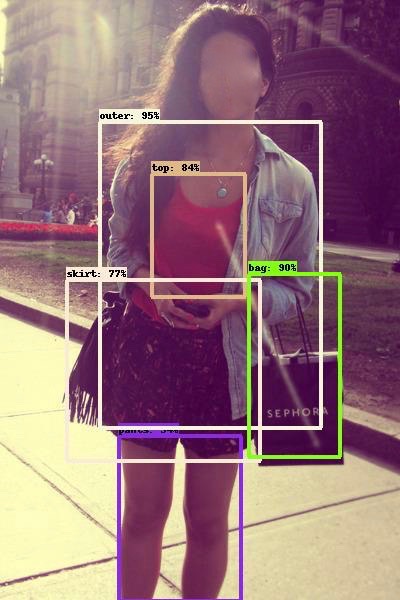} &
 \includegraphics[width=.11\textwidth]{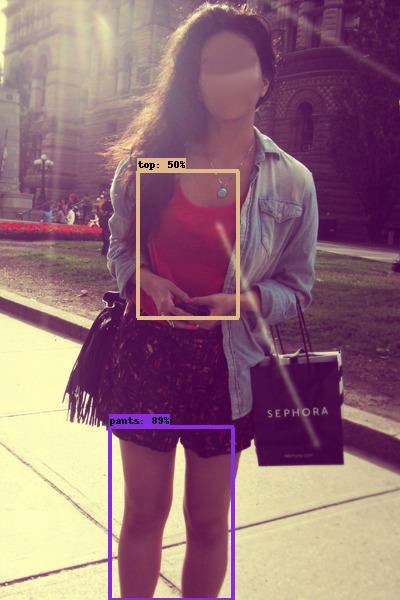}\\

 \includegraphics[width=.11\textwidth]{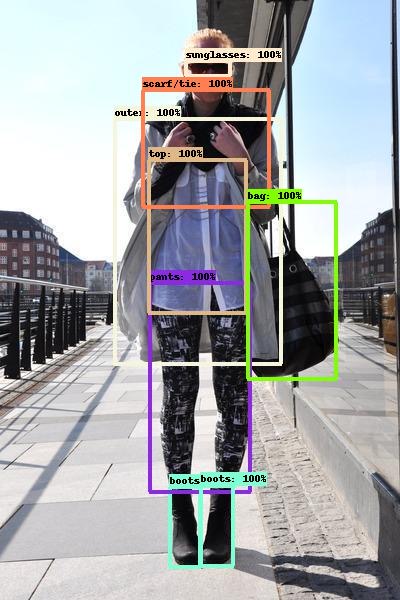} & 
 \includegraphics[width=.11\textwidth]{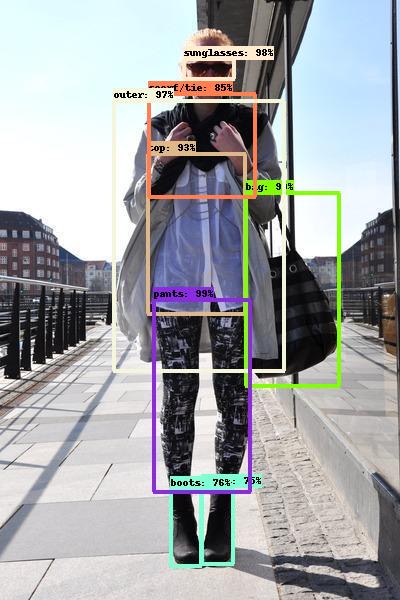} &
 \includegraphics[width=.11\textwidth]{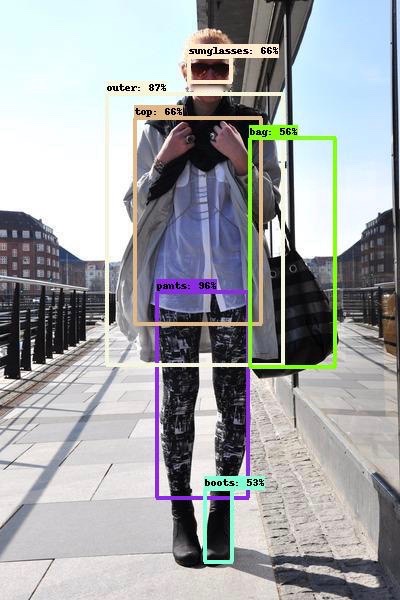} &
 \includegraphics[width=.11\textwidth]{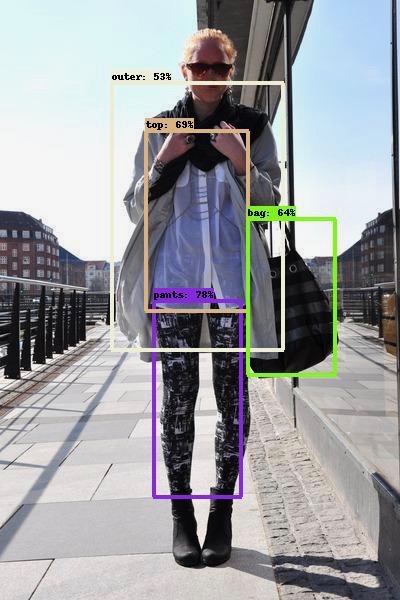}\\
 \end{tabular}
 \caption{Qualitative comparisons of Faster RCNN, SSD and YOLO.}
\label{fig:detection-qual} 
\end{figure}

In Figure~\ref{fig:map-per-cat}, we show the precision-recall curve per category of each detector with the IOU threshold $0.5$. 
Apparently, the Faster RCNN detector achieves the best overall accuracy and on all categories, which aligns well with our expectation. 
Specifically, all the detectors do not perform well on \emph{scarf\&tie} category while all achieve promising results (with mAP over $0.9$) on the \emph{pants}, \emph{headwear} and \emph{sunglasses} categories. Since there are abundant samples of pants in various poses, and sunglasses and headwear are usually of rigid shape without occlusion, it is easier for the model to recognize and localize these instances from street photos. In contrast, scarves and ties are highly deformable and often occluded by human bodies or other clothes, adding great difficulty to model training. 
Qualitative comparisons of the three detectors are presented in Figure~\ref{fig:detection-qual}.

\begin{figure*}[t]
\begin{center}
\includegraphics[width=.32\linewidth]{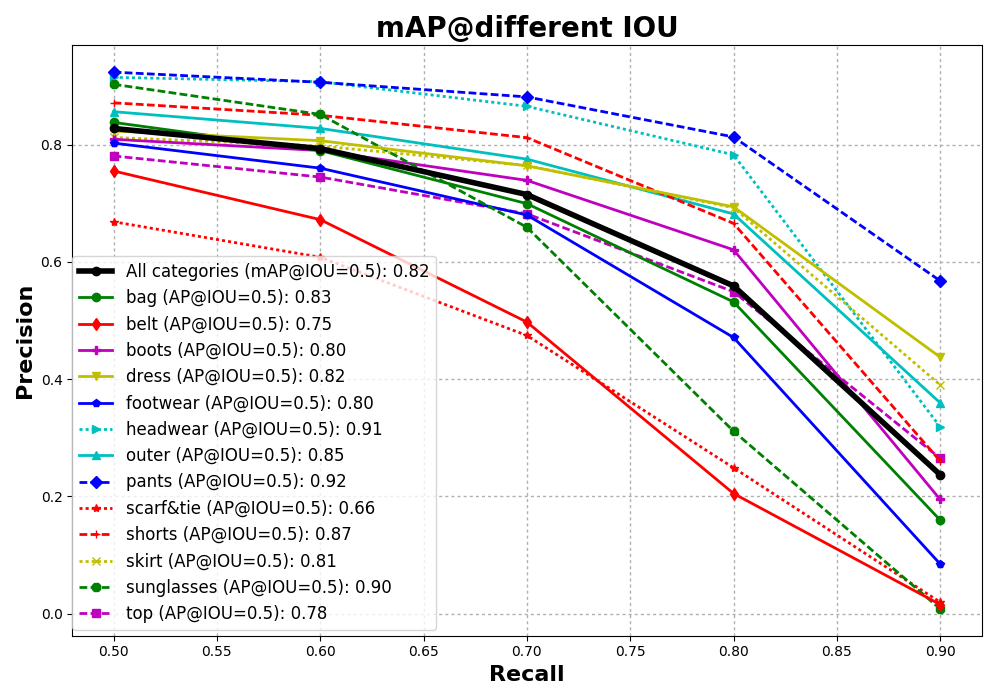}
\includegraphics[width=.32\linewidth]{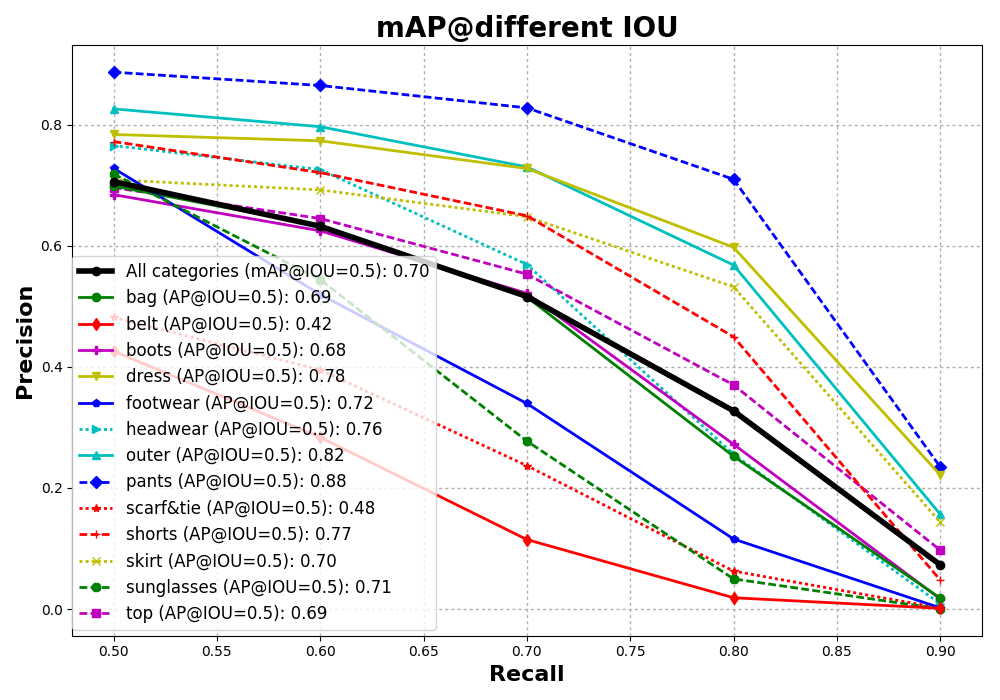}
\includegraphics[width=.32\linewidth]{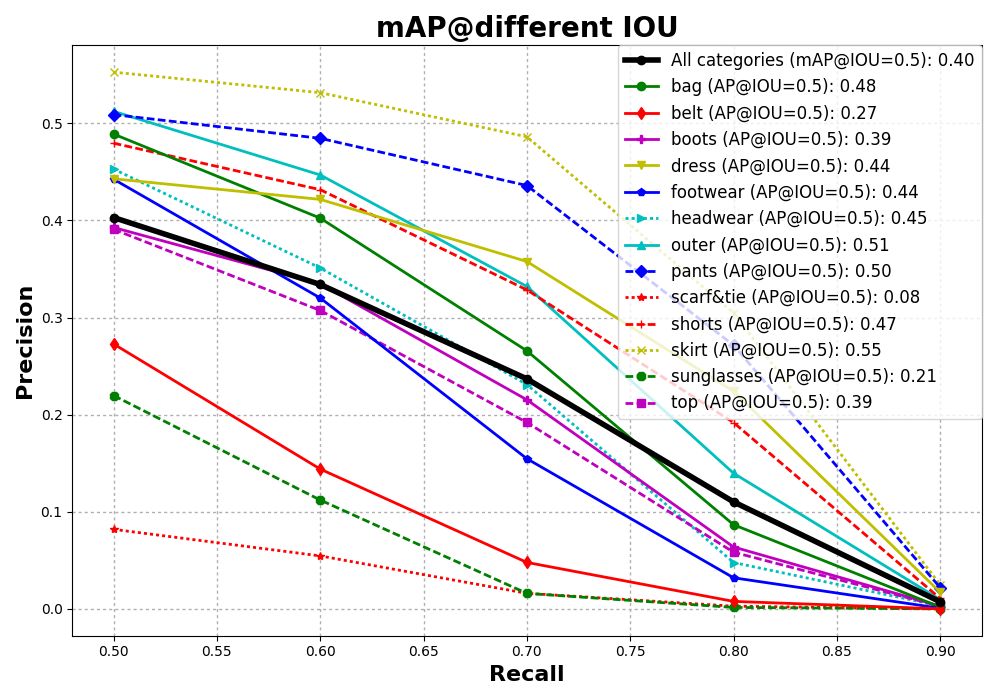}
\end{center}
\vspace{-2mm}
\caption{Performance comparison of Faster RCNN (left), SSD (middle) and YOLO (right). Best view in color.}
\label{fig:map-vs-iou}
\end{figure*}

To evaluate the robustness of these detectors against varying IOU threshold, we increase the threshold from $0.5$ to $0.9$ by $0.1$ increment and calculate mAP regarding to each IOU threshold. 
The results are shown in Figure~\ref{fig:map-vs-iou}. 
As expected, the mAP and AP per category decreases as more rigorous criterion is applied. 
We find that the detectors are relatively robust for categories such as \emph{pants} and \emph{headwear} as the AP only slightly drops, remaining over $0.8$ even though the IOU threshold is as high as $0.8$. It shows that these categories are easier to handle.
In contrast, categories of small and deformable objects like \emph{sunglasses}, \emph{belt} and \emph{scarf\&tie} are more difficult for the detectors. A stand-out example is \emph{sunglasses} category, on which the Faster RCNN achieves the second best AP when IOU=$0.5$. However, when we increase the IOU threshold, the corresponding AP dramatically drops to the worst when IOU=$0.9$. Such case implies that the detected bounding boxes do not have a high overlap ratio with the groundtruth. 
Therefore, more effort should be put into developing detectors that can better handle small and highly deformable objects for fashion.

\subsection{Semantic Segmentation for Fashion}
ModaNet dataset provides pixel-wise annotation to enable semantic segmentation research for fashion.

\subsubsection{Groundtruth generation}
Given the polygon annotations, we generate pixel-wise annotations. For the polygon annotations that cover a single object, it is straightforward to obtain the corresponding pixel-wise annotations. 
In other cases, the same object might be occluded and there are multiple polygon annotations to cover the same object. In the human annotations, this information is included and the multiple polygon annotations of the same object are grouped together. For objects like shoes, we assign the same semantic labels to the both left and right shoes as long as they are visible. We follow the same training/validation split as fashion item detection, so that we have $52,377$ images in the training set and $2,799$ images in the validation set.

\subsubsection{Approaches}
Semantic segmentation is a popular research direction. Many of the semantic segmentation approaches are developed to tackle generic things and stuff from images. We evaluate several most representative approaches on ModaNet and provide the baseline results to motivate future research. These approaches are: FCNs~\cite{Shelhamer/pami2016}\footnote{\url{https://github.com/shelhamer/fcn.berkeleyvision.org}}, CRFasRNN~\cite{Zheng/iccv2015}\footnote{\url{https://github.com/torrvision/crfasrnn}} and DeepLabv3+~\cite{Chen/cvpr2018}\footnote{\url{https://github.com/tensorflow/models/tree/master/research/deeplab}}. 
DeepLab~\cite{Chen/pami2018} is a state-of-the-art semantic image segmentation approach on the PASCAL VOC dataset. It has been improved by using atrous convolution, multi-scale training, and various base models including VGG, ResNet-101 and Xception. 
In order to measure the performance of different approaches, we adapt the standard mean pixel intersection-over-union (IOU), with the mean taken over all classes, as well as precision, recall, and F-1 scores. 

For FCN-32~\cite{Shelhamer/pami2016}, FCN-8~\cite{Shelhamer/pami2016}, and CRFasRNN~\cite{Zheng/iccv2015}, we follow the pipeline in Caffe from the original authors. 
We adapt the base model of VGG~\cite{Simonyan/iclr2015} network with batch normalization~\cite{simon2016cnnmodels}, which has obtained $26.9\%$ top-1 error on the ImageNet-2012 semantic segmentation task.
We train those models using batch size $1$, base learning rate $1\times~\mathrm{10}{-9}$, learning rate decay step $52,378$, step learning strategy, and stochastic gradient descent optimization method. 
We use the default set of initial parameters of CRFasRNN for PASCAL VOC dataset. We For FCN, we conduct experiments using Caffe, and evaluate its performance (using original VGG network) using Pytorch.

For DeepLabv3+~\cite{Chen/cvpr2018}, we take the public available TensorFlow implementation and ImageNet pre-trained Xception-65 model~\cite{xception/cvpr2017}, and fine-tune on the ModaNet dataset for $115$ epochs using batch size $6$, base learning rate $5\times10^{-4}$, learning rate decay step $52,378$, step learning strategy and stochastic gradient descent optimization method. 
The base model Xception-65 has obtained $21.0\%$ top-1 error on the ImageNet-2012 semantic segmentation task. 


\subsubsection{Performance comparison}

\begin{figure}[!htb]
\begin{center}
\includegraphics[width=1\linewidth]{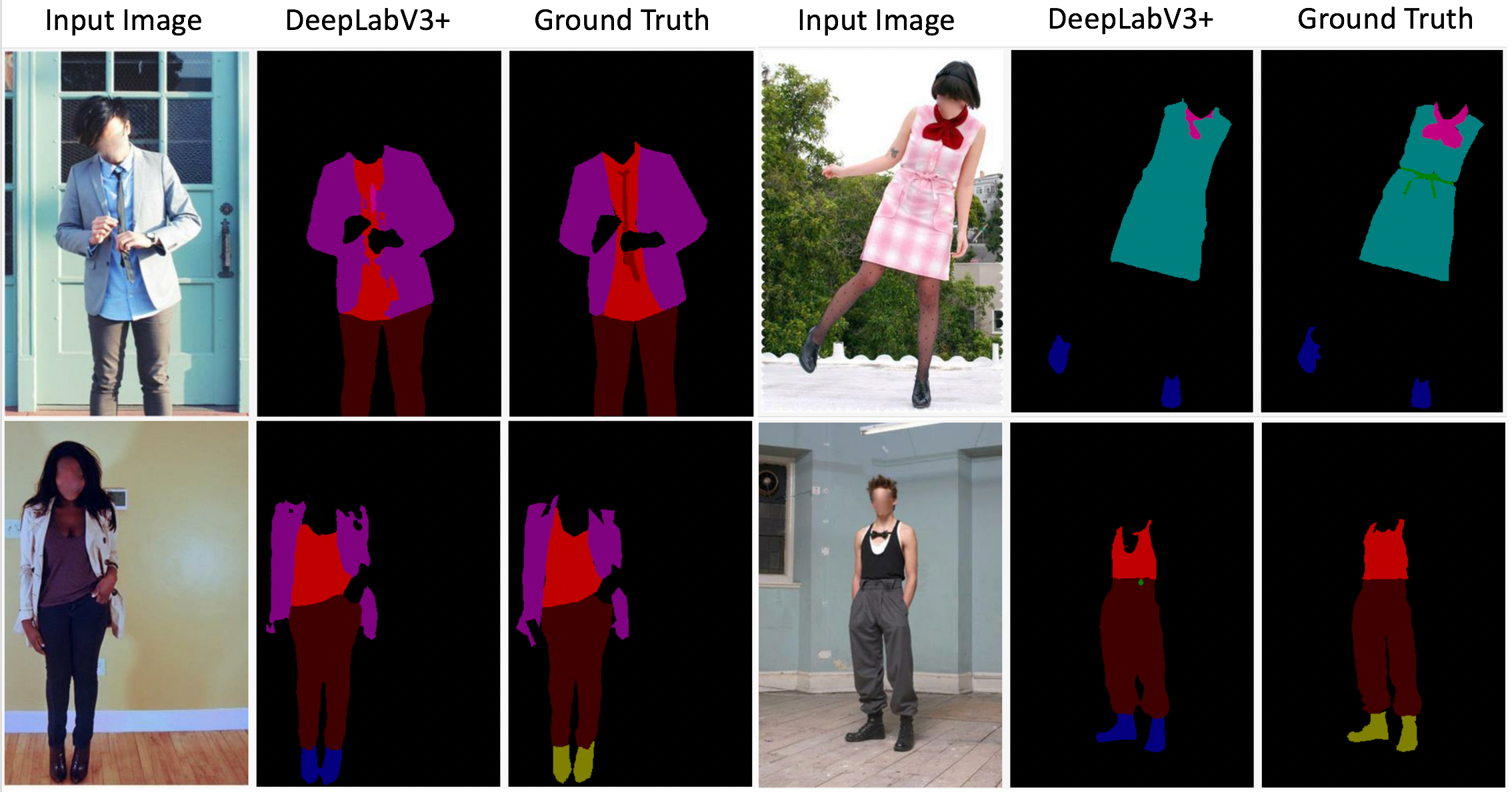}
\end{center}
\caption{Failure case. The 1st column is input, the 2nd column is result of DeeplabV3+, and the 3rd column is ground truth. DeeplabV3+ misses the ties in the 1st row and confused between boots and shoes in the 2nd row.}
\label{fig:failurecase}
\end{figure}

We find that DeepLabV3+ performs significantly better than the alternative approaches across all metrics, which is consistent with the results on PASCAL VOC leaderboard. 
As shown in Table~\ref{tab:iou-per-class}, Table~\ref{tab:precision-per-class}, Table~\ref{tab:recall-per-class}, and Table~\ref{tab:f1-per-class}, CRFasRNN improves performance for categories with large extend, while slightly decreases the performance for categories with smaller extend. 
Since the bilateral filter in CRFasRNN uses the same set of parameters for all categories, CRFasRNN tends to over-smooth out small objects, such as \emph{sunglasses}, resulting poorer performance.
CRFasRNN maintains better shape of some objects like \emph{outer} and \emph{pants}, since these objects are often color consistent. Figure~\ref{fig:failurecase} shows the failure cases.

\begin{figure}[t]
\begin{center}
\includegraphics[width=0.80\linewidth]{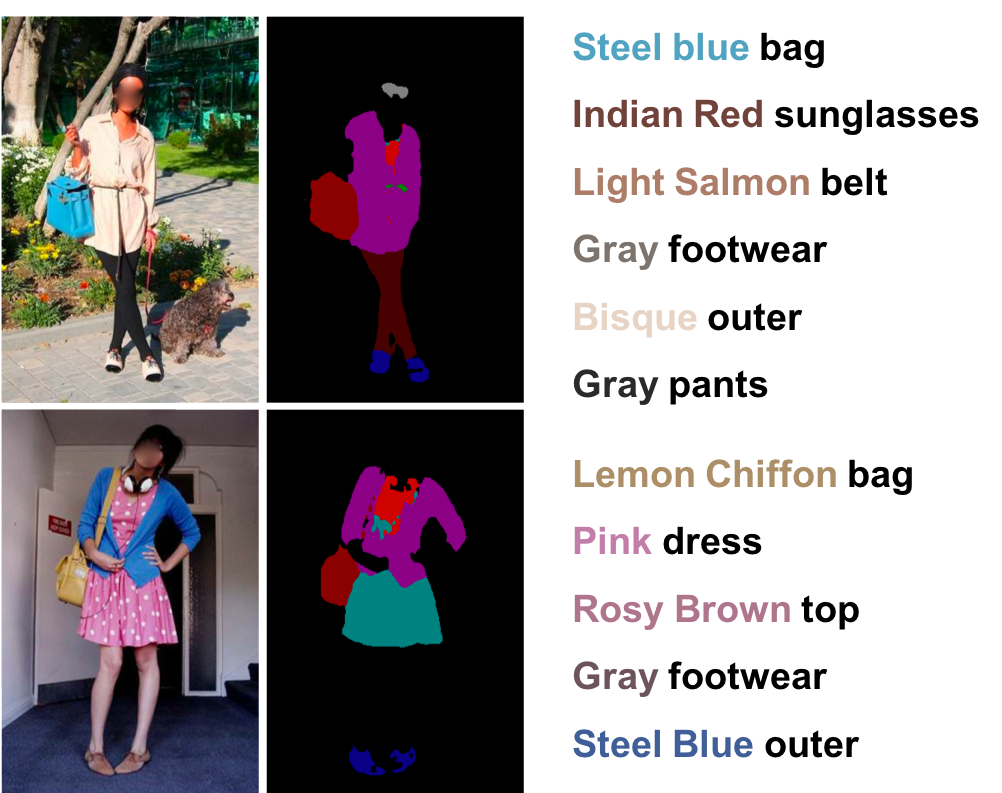}
\end{center}
\caption{Color attribute prediction using semantic image segmentation. The color names are printed with the predicted mean RGB values for the corresponding category. Best viewed in color.}
\label{fig:color_prediction}
\end{figure}


\begin{table*}[htb]
\small 
\centering
\caption{{IoU per category of evaluated semantic segmentation approaches.}}
\label{tab:iou-per-class}
\vspace{-2mm}
\begin{tabular}{|l|l|l|l|l|l|l|l|l|l|l|l|l|l|l|}
\hline
Method & bg & bag & belt  & boots & footwear & outer & dress & sunglasses & pants  & top & shorts  & skirts & headwear  & scarf\&tie \\ \hline
FCN-32~\cite{Shelhamer/pami2016} & 0.95 & 0.27 & 0.12 & 0.32 & 0.33 & 0.36 & 0.28 & 0.25 & 0.51 & 0.38 & 0.40 & 0.28 & 0.33 & 0.17      \\ \hline
FCN-16~\cite{Shelhamer/pami2016} & 0.96 & 0.26 & 0.19 & 0.32 & 0.38 & 0.35 & 0.25 & 0.37 & 0.51 & 0.38 & 0.40 & 0.23 & 0.41 & 0.16      \\ \hline
FCN-8~\cite{Shelhamer/pami2016} & 0.96 & 0.24 & 0.21 & 0.32 & 0.40 & 0.35 & 0.28 & 0.41 & 0.51 & 0.38 & 0.40 & 0.24 & 0.44 & 0.18      \\ \hline
FCN-8satonce~\cite{Shelhamer/pami2016} & 0.96 & 0.26 & 0.20 & 0.31 & 0.40 & 0.35 & 0.29 & 0.36 & 0.50 & 0.39 & 0.38 & 0.26 & 0.44 & 0.16      \\ \hline
CRFasRNN~\cite{Zheng/iccv2015} & 0.96 & 0.30 & 0.18 & 0.41 & 0.39 & 0.43 & 0.32 & 0.36 & 0.56 & 0.40 & 0.44 & 0.26 & 0.45 & 0.22      \\ \hline
DeepLabV3+~\cite{Chen/cvpr2018} & 0.98 & 0.42 & 0.28 & 0.40 & 0.51 & 0.56 & 0.52 & 0.46 & 0.68 & 0.55 & 0.53 & 0.41 & 0.55 & 0.31      \\ \hline
\end{tabular}
\end{table*}

\begin{table*}[htb]
\small
\centering
\caption{{Precision per category of evaluated semantic segmentation approaches}.}
\label{tab:precision-per-class}
\vspace{-2mm}
\begin{tabular}{|l|l|l|l|l|l|l|l|l|l|l|l|l|l|l|}
\hline
Method & bg & bag & belt  & boots & footwear & outer & dress & sunglasses & pants  & top & shorts  & skirts & headwear  & scarf\&tie \\ \hline
FCN-32~\cite{Shelhamer/pami2016} & 0.97 & 0.52 & 0.37 & 0.59 & 0.51 & 0.53 & 0.44 & 0.53 & 0.60 & 0.53 & 0.56 & 0.45 & 0.47 & 0.46      \\ \hline
FCN-16~\cite{Shelhamer/pami2016} & 0.97 & 0.52 & 0.42 & 0.64 & 0.58 & 0.47 & 0.38 & 0.66 & 0.64 & 0.59 & 0.58 & 0.35 & 0.65 & 0.36      \\ \hline
FCN-8~\cite{Shelhamer/pami2016} & 0.97 & 0.42 & 0.51 & 0.66 & 0.58 & 0.52 & 0.46 & 0.74 & 0.61 & 0.49 & 0.62 & 0.45 & 0.74 & 0.53      \\ \hline
FCN-8satonce~\cite{Shelhamer/pami2016} & 0.97 & 0.51 & 0.43 & 0.75 & 0.59 & 0.52 & 0.45 & 0.74 & 0.59 & 0.52 & 0.51 & 0.44 & 0.72 & 0.55      \\ \hline
CRFasRNN~\cite{Zheng/iccv2015} & 0.96 & 0.49 & 0.57 & 0.66 & 0.68 & 0.58 & 0.58 & 0.65 & 0.76 & 0.61 & 0.61 & 0.37 & 0.73 & 0.44      \\ \hline
DeepLabV3+~\cite{Chen/cvpr2018} & 0.99 & 0.62 & 0.53 & 0.75 & 0.62 & 0.70 & 0.67 & 0.74 & 0.75 & 0.69 & 0.69 & 0.56 & 0.74 & 0.51      \\ \hline
\end{tabular}
\end{table*}

\begin{table*}[htb]
\small
\centering
\caption{{Recall per category of evaluated semantic segmentation approaches}.}
\label{tab:recall-per-class}
\vspace{-2mm}
\begin{tabular}{|l|l|l|l|l|l|l|l|l|l|l|l|l|l|l|}
\hline
Method & bg & bag & belt  & boots & footwear & outer & dress & sunglasses & pants  & top & shorts  & skirts & headwear  & scarf\&tie \\ \hline
FCN-32~\cite{Shelhamer/pami2016} & 0.98 & 0.43 & 0.23 & 0.48 & 0.53 & 0.62 & 0.60 & 0.44 & 0.81 & 0.63 & 0.66 & 0.53 & 0.67 & 0.31   \\ \hline
FCN-16~\cite{Shelhamer/pami2016} & 0.98 & 0.42 & 0.37 & 0.47 & 0.60 & 0.69 & 0.61 & 0.54 & 0.78 & 0.57 & 0.64 & 0.59 & 0.62 & 0.36   \\ \hline
FCN-8~\cite{Shelhamer/pami2016} & 0.99 & 0.47 & 0.37 & 0.44 & 0.64 & 0.63 & 0.57 & 0.53 & 0.81 & 0.71 & 0.61 & 0.47 & 0.57 & 0.28   \\ \hline
FCN-8satonce~\cite{Shelhamer/pami2016} & 0.99 & 0.42 & 0.38 & 0.39 & 0.63 & 0.63 & 0.58 & 0.48 & 0.83 & 0.68 & 0.69 & 0.51 & 0.59 & 0.24   \\ \hline
CRFasRNN~\cite{Zheng/iccv2015} & 0.99 & 0.53 & 0.26 & 0.59 & 0.52 & 0.70 & 0.50 & 0.54 & 0.71 & 0.58 & 0.66 & 0.62 & 0.60 & 0.41     \\ \hline
DeepLabV3+~\cite{Chen/cvpr2018} & 0.99 & 0.61 & 0.48 & 0.50 & 0.78 & 0.78 & 0.73 & 0.60 & 0.88 & 0.74 & 0.74 & 0.69 & 0.72 & 0.54   \\ \hline
\end{tabular}
\end{table*}

\begin{table*}[htb]
\small
\centering
\caption{{F-1 score per category of evaluated semantic segmentation approaches}.}
\label{tab:f1-per-class}
\vspace{-2mm}
\begin{tabular}{|l|l|l|l|l|l|l|l|l|l|l|l|l|l|l|}
\hline
Method & bg & bag & belt  & boots & footwear & outer & dress & sunglasses & pants  & top & shorts  & skirts & headwear  & scarf\&tie \\ \hline
FCN-32~\cite{Shelhamer/pami2016} & 0.98 & 0.58 & 0.43 & 0.67 & 0.58 & 0.67 & 0.70 & 0.53 & 0.81 & 0.67 & 0.79 & 0.74 & 0.65 & 0.50   \\ \hline
FCN-16~\cite{Shelhamer/pami2016} & 0.98 & 0.57 & 0.50 & 0.65 & 0.66 & 0.69 & 0.68 & 0.65 & 0.82 & 0.66 & 0.79 & 0.72 & 0.71 & 0.52   \\ \hline
FCN-8~\cite{Shelhamer/pami2016} & 0.98 & 0.56 & 0.52 & 0.65 & 0.69 & 0.67 & 0.68 & 0.70 & 0.83 & 0.67 & 0.78 & 0.67 & 0.71 & 0.51   \\ \hline
FCN-8satonce~\cite{Shelhamer/pami2016} & 0.98 & 0.57 & 0.52 & 0.65 & 0.69 & 0.68 & 0.69 & 0.65 & 0.84 & 0.68 & 0.80 & 0.70 & 0.72 & 0.47   \\ \hline
CRFasRNN~\cite{Zheng/iccv2015} & 0.98 & 0.62 & 0.48 & 0.74 & 0.63 & 0.74 & 0.68 & 0.63 & 0.83 & 0.69 & 0.82 & 0.77 & 0.73 & 0.60      \\ \hline
DeepLabV3+~\cite{Chen/cvpr2018} & 0.99 & 0.72 & 0.59 & 0.72 & 0.78 & 0.82 & 0.84 & 0.74 & 0.90 & 0.78 & 0.85 & 0.85 & 0.81 & 0.70   \\ \hline
\end{tabular}
\end{table*}

\subsection{Polygon Prediction for Fashion}
Recognizing clothing design and detecting polygon keypoints of fashion items are useful in fashion applications.
Polygon is another typical form of annotations in semantic segmentation, apart from pixel-level segmentation masks. 
It is considered to be more accurate than the alternatives such as superpixel-based annotations, although manually annotating images using polygons is time-expensive.
In this work, given high-quality polygons from human annotators, we conduct experiments to predict polygons for individual fashion items given an input image.

\subsubsection{Approaches}
Recent approaches such as PolygonRNN~\cite{Castrejon/cvpr2017} and Polygon-RNN++~\cite{Acuna/cvpr2018} have addressed the problem of polygon prediction directly using neural networks. We set up the baseline performance on ModaNet using the pre-trained model of Polygon-RNN++. 
This model is an encoder-decoder network. The encoder produces image features that are used to predict the first vertex, and then the first vertex and the image features are fed to the recurrent decoder. 
The recurrent neural network exploits the visual attention at each time step to produce polygon vertices. 
A learned evaluator is employed to select the best polygon from a set of candidates proposed by the decoder. 
In the final stage, a graph-based neural network re-adjusts the polygons and augments them with additional vertices at a higher resolution. 
The base model in the encoder is a modified ResNet-50, which has reduced stride and dilation factors. 

\subsubsection{Performance}
In our experiment, we adapt the public available Polygon-RNN++ model that is pre-trained on the Cityscape dataset to produce the baseline performance. 
We form the inputs of Polygon-RNN++ using the cropped images based on the Faster-RCNN detection results. The polygon predictions of polygon-RNN++ are evaluated on the validation set of ModaNet dataset. We convert the polygon predictions to the mask-like predictions.
A perfect polygon prediction should give the same semantic image segmentation mask created by human annotators. 
Using the masks for semantic segmentation as groundtruth, we evaluate how well the results from the pre-trained polygon-RNN++ model align well with the segmentation masks. The Polygon-RNN++ model achieves mean IOU $30.7\%$, mean precision $83.4\%$, mean recall $32.5\%$ and mean F-1 score $45.0\%$. 
We hope such preliminary baseline results on the ModaNet dataset would motivate future research on polygon prediction.

\subsection{Color Attribute Prediction Prototype}
One application of semantic segmentation is to predict the color attribute name given a fashion product. 
We develop such a prototype based on ModaNet dataset. We first conduct semantic segmentation, and then predict the color attribute names by mapping the mean RGB values for each segment to a fine-grained color name space\footnote{\url{https://github.com/ayushoriginal/Optimized-RGB-To-ColorName}}. The example results are presented in Figure~\ref{fig:color_prediction}. The image and the predicted masks are shown on the left, and the predicted text including the color attribute names and the fashion object category names is shown on the right. 

\section{Conclusions}
We introduce a new large-scale image dataset with polygon annotations. This dataset contains street fashion photos that exhibit various challenges, such as human poses, lighting, occlusion, deformation, \etc, which well captures the scenarios in the real world.
Different types of annotations are provided including polygons, pixel-level segmentation masks, and bounding boxes. 
We have conducted experiments on object detection, semantic segmentation, and polygon predictions. We also demonstrate that the proposed dataset is suitable for training models to achieve promising application such as color attribute prediction. 

\section*{Acknowledgement}
The authors thank Kota Yamaguchi for providing the Paperdoll dataset.

\bibliographystyle{ACM-Reference-Format}
\bibliography{bibs}

\end{document}